\pdfoutput=1

\documentclass[11pt]{article}

\usepackage{amsmath}
\usepackage{wrapfig}
\usepackage{amsfonts}
\usepackage{dsfont}
\usepackage[table]{xcolor}
\usepackage{multirow}
\usepackage{booktabs}
\usepackage{subcaption}
\usepackage{etoolbox}
\usepackage{siunitx}
\usepackage{bbm}
\usepackage{amssymb}
\usepackage{lipsum}
\usepackage[most]{tcolorbox}
\usepackage{xcolor}
\usepackage{colortbl}
\usepackage{algorithm2e}
\usepackage{multicol}
\usepackage{algpseudocode}
\usepackage{svg}
\usepackage{bbding}
\usepackage[english]{babel}
\usepackage{tikz}
\usetikzlibrary{shapes, arrows.meta, positioning}
\definecolor{myorange}{HTML}{fa8072}
\definecolor{mygreen}{HTML}{9DC183}
\usepackage{enumitem}

\newcommand{\multirowcell}[1]{\begin{tabular}[c]{@{}c@{}}#1\end{tabular}}

\usepackage[final]{acl}

\usepackage{times}
\usepackage{latexsym}

\usepackage[T1]{fontenc}

\usepackage[utf8]{inputenc}

\usepackage{microtype}

\usepackage{inconsolata}

\usepackage{graphicx}

\usepackage{todonotes}
\usepackage{marginnote}
\usepackage{forest}

\usepackage{booktabs} 
\usepackage[table]{xcolor}
\usepackage{placeins}

\newcommand{\retr}{\textbf{r}}

\usepackage{amsmath}
\usepackage{amssymb}
\usepackage{mathtools}
\usepackage{amsthm}
\usepackage{easyeqn}
\usepackage{natbib}
\usepackage{hyperref}
\usepackage{ulem}
\usepackage{bm}
\usepackage{multirow}
\usepackage{xspace}
\usepackage{float}

\usepackage[capitalize,noabbrev]{cleveref}

\theoremstyle{plain}

\theoremstyle{definition}

\theoremstyle{remark}

\newcommand{\xv}{\mathbf{x}}
\newcommand{\yv}{\mathbf{y}}

\newcommand{\RR}{\mathbb{R}}

\newcommand{\DC}{\mathcal{D}}

\newcommand{\FC}{\mathcal{F}}

\newcommand{\SC}{\mathcal{S}}

\newcommand{\llama}{Llama 3B Instruct\xspace}
\newcommand{\llamalarge}{Llama 8B Instruct\xspace}
\newcommand{\falcon}{Falcon 3B Base\xspace}
\newcommand{\gemma}{Gemma 12B Instruct\xspace}
\newcommand{\gemmasmall}{Gemma 4B Instruct\xspace}
\newcommand{\gpto}{GPT-4o\xspace}
\newcommand{\gpt}{GPT-4\xspace}

\newcommand\freefootnote[1]{%
  \let\thefootnote\relax%
  \footnotetext{#1}%
  \let\thefootnote\svthefootnote%
}

\title{Faithfulness-Aware Uncertainty Quantification for \\ Fact-Checking the Output of Retrieval-Augmented Generation}

\author{
  Ekaterina Fadeeva\textsuperscript{1 $\diamondsuit$} \quad Aleksandr Rubashevskii\textsuperscript{2 $\diamondsuit$} \quad Dzianis Piatrashyn\textsuperscript{2} \\
  \textbf{Roman Vashurin\textsuperscript{2} ~~ Shehzaad Dhuliawala\textsuperscript{1} ~~ Artem Shelmanov\textsuperscript{2} ~~ Timothy Baldwin\textsuperscript{2}} \\
  \textbf{Preslav Nakov\textsuperscript{2} \quad Mrinmaya Sachan\textsuperscript{1} \quad Maxim Panov\textsuperscript{2}} \\
  \textsuperscript{1}ETH Zürich \quad \textsuperscript{2}MBZUAI
  \\
  \href{mailto:efadeeva@ethz.ch}{\{efadeeva,sdhuliawala,msachan\}@ethz.ch} \\
  \href{mailto:aleksandr.rubashevskii@mbzuai.ac.ae}{\{aleksandr.rubashevskii,preslav.nakov,maxim.panov\}@mbzuai.ac.ae} 
}

\newcommand{\franq}{\textsc{franq}\xspace}

\begin{document}
\definecolor{TodoColor}{rgb}{1,0.7,0.6}
\definecolor{aquamarine}{rgb}{0.5, 1.0, 0.83}
\definecolor{some_blue}{rgb}{0.4, 0.4, 1.0}
\newcommand{\todonote}[3][]{\todo[color=#2,size=\scriptsize,fancyline,caption={},#1]{#3}}
\newcommand{\todox}[2][]{\todonote[#1]{TodoColor}{\textbf{TODO:} #2}}
\newcommand{\vilem}[2][]{\todonote[#1]{pink}{\textbf{Vilém:} #2}}
\newcommand{\Vilem}[2][]{\vilem[inline,#1]{#2}}
\newcommand{\szd}[2][]{\todonote[#1]{teal}{\textbf{shehz:} #2}}
\newcommand{\Peng}[2][]{\peng[inline,#1]{#2}}
\newcommand{\mrinmaya}[2][]{\todonote[#1]{pink}{\textbf{Mrinmaya:} #2}}
\newcommand{\rediska}[2][]{\todonote[#1]{aquamarine}{\textbf{rediska:} #2}}
\newcommand{\maxim}[2][]{\todonote[#1]{some_blue}{\textbf{Maxim:} #2}}
\newcommand{\Mrinmaya}[2][]{\mrinmaya[inline,#1]{#2}}

\newcommand{\TODO}[2][]{\todox[inline,#1]{#2}}
\newcommand{\TODOMARK}{\textcolor{black}{\sethlcolor{TodoColor} \small \hl{\textbf{TODO}}}\xspace}
\newcommand{\CITEME}{\textcolor{black}{\small \hl{\textbf{CITEME}}}\xspace}

\newcommand\comet[2][]{Comet$^{#2}_\textrm{#1}$\xspace}
\newcommand{\hrefEmail}[2]{\href{mailto:#1}{\color{black}{#2}}}
\newcommand{\whitezero}{\textcolor{white}{0}}
\newcommand{\offsetminus}{\hspace{-1.2mm}-}

\newcommand{\hlc}[2][yellow]{{%
    \colorlet{foo}{#1}%
    \sethlcolor{foo}\hl{#2}}%
}
\definecolor{FindingsColor}{gray}{0.85}
\newcommand{\hlfinding}[1]{\hlc[FindingsColor]{#1}}

\makeatletter\def\Hy@Warning#1{}\makeatother
\let\svthefootnote\thefootnote
\newcommand\blankfootnote[1]{%
  \let\thefootnote\relax\footnotetext{#1}%
  \let\thefootnote\svthefootnote%
}


\newcommand{\legend}[3]{
\null\hspace{3mm}
\makebox[22mm][l]{
    \textcolor[HTML]{#1}{
    \rule[3pt]{10pt}{1.5pt}
    \hspace{-13pt}
    \raisebox{0.5pt}{\scalebox{1.5}{$\bullet$}}}
    #2
}
\makebox[8mm][l]{#3}
}

\newcommand{\legendShort}[2]{
\null\hspace{1mm}
\makebox[21mm][l]{
    \textcolor[HTML]{#1}{
    \rule[3pt]{10pt}{1.5pt}
    \hspace{-13pt}
    \raisebox{0.5pt}{\scalebox{1.5}{$\bullet$}}}
    #2
}
}

\newcommand\anonymized{\texttt{\bf[anonymized]}\xspace}

  \maketitle

  \begin{abstract}
    Large Language Models (LLMs) enhanced with retrieval, an approach known as Retrieval-Augmented Generation (RAG), have achieved strong performance in open-domain question answering. However, RAG remains prone to hallucinations: factually incorrect outputs may arise from inaccuracies in the model's internal knowledge and the retrieved context. Existing approaches to mitigating hallucinations often conflate factuality with faithfulness to the retrieved evidence, incorrectly labeling factually correct statements as hallucinations if they are not explicitly supported by the retrieval.
    In this paper, we introduce \franq, a new method for hallucination detection in RAG outputs. \franq applies distinct uncertainty quantification (UQ) techniques to estimate factuality, conditioning on whether a statement is faithful to the retrieved context. To evaluate \franq and competing UQ methods, we construct a new long-form question answering dataset annotated for both factuality and faithfulness, combining automated labeling with manual validation of challenging cases.
    Extensive experiments across multiple datasets, tasks, and LLMs show that \franq achieves more accurate detection of factual errors in RAG-generated responses compared to existing approaches.
    Our implementation is available at \url{https://github.com/stat-ml/rag_uncertainty}.
  \end{abstract}


\section{Introduction}

\freefootnote{$\diamondsuit$ Equal contribution}

  Large Language Models (LLMs) are increasingly employed across a wide range of tasks, including natural language understanding, generation, and reasoning. However, LLMs are prone to generating plausible but factually incorrect generations, a phenomenon known as \emph{hallucination}, arising from factors such as insufficient training data coverage, input ambiguity, as well as architectural constraints~\cite{huang2025survey}.
  
  Retrieval-Augmented Generation (RAG; \citealp{lewis2020retrieval}) mitigates this issue by incorporating dynamically retrieved external knowledge into the generation process, which can partially mitigate factual inaccuracies~\cite{shuster2021retrieval}.

  However, RAG systems still produce hallucinations~\cite{shi2023large}.
  The use of retrieved information complicates both their detection and source attribution, as models become more confident in generating statements that appear in the retrieval, regardless of factual correctness~\citep{kim2024speak}. At the same time, the retrieved passages themselves may be erroneous, incomplete, or completely irrelevant with respect to the query~\citep{shi2023large, ding2024retrieve}. Conversely, even when retrieval is accurate, inconsistencies can emerge between the model's internal knowledge and the retrieved data~\citep{wang2024astute, wang2025retrievalaugmented}.


  Thus, an important question is how to define \textit{hallucination} in RAG, given the interplay between the model's internal knowledge and the retrieved context. One approach considers any content not supported by the retrieved context as hallucination~\cite{wu2023ragtruth}. However, we argue that hallucination should instead be defined based on factual inaccuracies: statements outside the retrieved context should not be considered hallucinations if they are factually correct.
  
  \begin{figure*}[htbp]
    \centering
    \includegraphics[width=\linewidth]{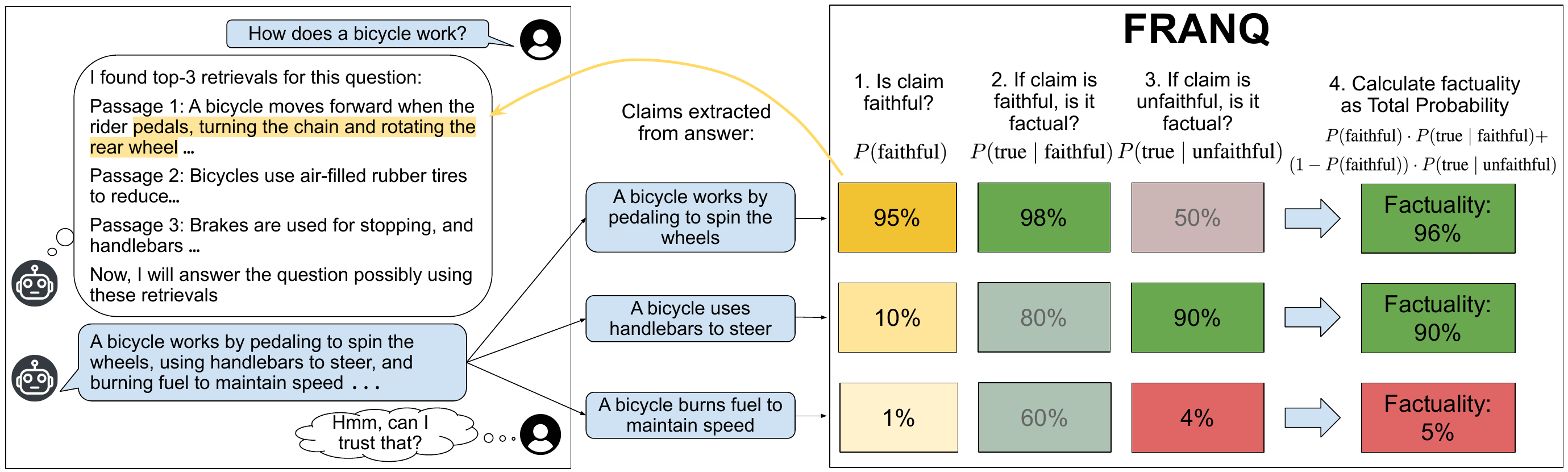}
    \caption{\franq illustration. \textit{Left}: A user poses a question, and the RAG retrieves relevant documents and formulates an answer, potentially using information from the retrieved documents. \textit{Middle}: The RAG output is decomposed into atomic claims. \textit{Right}: The \franq method assesses factuality by evaluating three components: (1)~faithfulness, (2)~factuality under faithful condition, and (3)~factuality under unfaithful condition. }
    \label{fig:our_method}
  \end{figure*}

  To address this distinction, we differentiate between \textit{factuality} and \textit{faithfulness}. Faithfulness refers to whether the generated output is semantically entailed by the retrieved context, while factuality indicates whether the content is objectively correct~\cite{maynez-etal-2020-faithfulness, dziri2022faithdial, yang2024factuality}. For RAG fact-checking, detecting non-factual claims is more critical than identifying unfaithful ones. 
  This distinction disentangles two core RAG failure modes: (\textit{i})~hallucinations caused by erroneous grounding in the retrieved context, and (\textit{ii})~factual errors stemming from the model's internal knowledge~\cite{rag_trustworthiness}.

  Here, we investigate the detection of non-factual statements produced by RAG using Uncertainty Quantification (UQ) techniques. We introduce \franq (\textbf{F}aithfulness-aware \textbf{R}etrieval \textbf{A}ugmented U\textbf{N}certainty \textbf{Q}uantification), a novel method that first evaluates the faithfulness of the generated response and subsequently applies different UQ methods based on the outcome. With this separation, \franq tailors its strategy to the specific RAG failure mode: whether it originates from retrieval grounding or from the model's own knowledge.
  
  We evaluate \franq on both long- and short-form question answering (QA) tasks. For long-form QA, where answers include multiple claims, we assess factuality at the claim level and introduce a new dataset with factuality annotations, combining automated labeling with manual validation. For short-form QA, we test our method on four QA datasets and treat each response as a single claim. 
  
  Our key \textbf{contributions} are as follows:
  \begin{itemize}
    \item We propose \franq, a UQ method for RAG that first assesses faithfulness and then applies different uncertainty estimation methods to faithful and unfaithful outputs (Section~\ref{sec:method}).

    \item We introduce a long-form QA factuality dataset for RAG with both factuality and faithfulness labels, built through automatic annotation and manual validation of difficult cases (Section~\ref{sec:rag_datasets}).

    \item We conduct extensive experiments on long- and short-form QA across several LLMs, showing that \franq improves factual error detection in RAG outputs over existing approaches (Section~\ref{sec:experiments}).
  \end{itemize}


\section{Uncertainty Quantification for RAG}
\label{sec:method}
  Let $\xv$ be the user query to the RAG system. The RAG system then retrieves $k$ passages denoted by $\retr = \{r_1, \dots, r_k\}$, from an external knowledge source using $\xv$ as the query, and uses an LLM to generate output $\yv$, conditioned on both $\xv$ and $\retr$.

  Autoregressive LLMs produce text sequentially, generating one token at a time.
  At each step $t$, the model samples a token $y_t \sim p(\cdot \mid \yv_{<t}, \xv, \retr)$, where $\yv_{<t}$ denotes the sequence of previously generated tokens.
  In the case of greedy decoding, this token is selected as the most likely outcome, i.e., $y_t = \arg\max_y{p(y \mid \yv_{<t}, \xv, \retr)}$.
  From $\yv$, we extract a set of $l$ atomic claims denoted as ${c_1, \dots, c_l}$. Each claim $c_i$ is associated with a specific span of tokens, $\SC(c_i)$, which represents the indices of the tokens in $\yv$ that correspond to this particular claim.
  
  A claim $c$ is considered \textit{factually true} if it is supported by established, verifiable knowledge, and \textit{factually false} otherwise. A claim is deemed \textit{faithful} with respect to the retrieved documents $\retr$, if it is entailed by them, and \textit{unfaithful} otherwise. Importantly, factuality and faithfulness capture different aspects of correctness: a claim may be factually true yet not grounded in the retrieved context, or faithful to the retrieval while still being factually incorrect. While most current benchmarks for evaluating RAG outputs focus on evaluating faithfulness~\cite{dziri2022faithdial,wu2023ragtruth}, our main objective is to assess the \textit{factuality} of claims.

\paragraph{General baselines.}
  A straightforward approach to hallucination detection is to apply standard UQ methods to LLM outputs conditioned on the joint prompt $(\xv, \retr)$. However, this ignores the structural asymmetry between $\xv$ and $\retr$.
  
  As an illustrative example, a common UQ baseline is to estimate the negative log-probability of a claim $c$ under the model distribution:
  \begin{equation}
    U(c \mid \xv, \retr) = -\sum_{t \in \SC(c)} \log p(y_t \mid \xv, \retr, \yv_{<t}).
    \label{eq:claim_prob}
  \end{equation}
  Table~\ref{tab:lmpoly_methods} summarizes several other UQ methods that can be applied in this general baseline setting.

\subsection{Faithfulness-aware Retrieval Augmented uNcertainty Quantification (\franq)}
  We introduce \franq, a new approach for assessing the factuality of claims in RAG outputs by leveraging UQ and explicitly treating $\xv$ and $\retr$ as separate inputs. The key idea is to first assess whether a generated claim is faithful to $\retr$ and then apply different UQ methods depending on the outcome. This yields the following decomposition of the probability that a claim $c$ is true:
  \begin{align}
    &P(c \text{ is true}) =  \label{eq:cuq} \\
    &P(c \text{ is faithful to \retr}) \cdot P(c \text{ is true} \mid \text{faithful}) + \nonumber \\
     &P(c \text{ is unfaithful to \retr}) \cdot P(c \text{ is true} \mid \text{unfaithful}), \nonumber
  \end{align}
  where $P(c \text{ is unfaithful to \retr})$ is calculated as $1 - P(c \text{ faithful to \retr})$. This decomposition isolates three probability components, each of which we approximate using specialized techniques described in Section~\ref{sec:franq_components}:
  \begin{enumerate}[itemsep=0pt, parsep=0pt, topsep=2pt]
    \item $P(c \text{ is faithful to \retr})$;

    \item $P(c \text{ is true} \mid \text{faithful})$;

    \item $P(c \text{ is true} \mid \text{unfaithful})$.
  \end{enumerate}
  An overview of \franq is visually depicted in Figure~\ref{fig:our_method}, and illustrative examples applied to individual claims are provided in Appendix~\ref{sec:franq_examples}.

\subsection{\franq Components}
\label{sec:franq_components}
  We now describe the components of equation~\eqref{eq:cuq}.

\paragraph{Faithfulness.}
  \label{sec:p_faith}
  To determine the degree to which a claim $c_i$ is entailed by the retrieved evidence $\retr$, we use \textit{AlignScore}, a RoBERTa-based similarity metric fine-tuned for factual alignment~\cite{zha2023alignscoreevaluatingfactualconsistency}. \textit{AlignScore} is specifically designed to measure factual consistency between a claim and context evidence, making it well suited for claim-level faithfulness estimation in RAG. Importantly, \textit{AlignScore} yields well-calibrated continuous faithfulness estimates rather than near-binary decisions; in practice, many claims exhibit intermediate values due to partial or implicit grounding. We analyze the distribution, calibration, and alternative faithfulness estimators in Appendix~\ref{section:appendix_faithfulness}.

  In long-form QA, we apply \textit{AlignScore} to each claim–retrieval pair $(c_i, \retr)$ to get the faithfulness estimate for claim $c_i$. In short-form QA, the answer $\yv$ is treated as a single claim, and we prepend the question context and evaluate \textit{AlignScore} on $(\xv \circ \yv, \retr)$, with `$\circ$' denoting string concatenation.

\paragraph{Factuality under unfaithful condition.}
  When a claim $c$ is unfaithful (not entailed by $\retr$), it originates from the LLM's internal knowledge. In this case, we estimate factuality using the model's probability estimates, avoiding distributional shifts arising from conditioning on retrieved context $\retr$. Specifically, we introduce a \textit{Parametric Knowledge} method, which computes the likelihood of $c$ based solely on the LLM's parametric knowledge~\citep{mallen2022not} without the retrieved evidence $\retr$:
  \begin{equation}
    p(c \mid \xv) = \prod_{t \in \SC(c)} p(y_t \mid \xv, \yv_{<t}).
    \label{eq:param_know}
  \end{equation}
  
  This method does not require generating new responses; instead, it reuses the original tokens and performs a forward pass through the LLM with the retrieved evidence removed from the input.
  
  In long-form QA, \textit{Parametric Knowledge} offers an effective estimate of factuality for unfaithful claims (see Section~\ref{sec:ablation_franq_components}). In short-form QA, a broader range of general UQ baselines is applicable, including methods based on sample diversity (see Table~\ref{tab:lmpoly_methods}). In this setting, the \textit{Sum of Eigenvalues}~\cite{lin2023generating} offers a better approximation of factuality (see Section~\ref{sec:ablation_franq_components}).
  Thus, we use \textit{Parametric Knowledge} for long-form QA and \textit{Sum of Eigenvalues} for short-form QA.

  \begin{table}[t]
\centering
\small

\resizebox{0.48\textwidth}{!}{

\begin{tabular}{cl|cc}

\toprule
\multirow{2}{*}{\textbf{Category}} & \multirow{2}{*}{\textbf{Uncertainty Quantification Method}}  & \multicolumn{2}{c}{\textbf{Suitable for}} \\
 & & \multirowcell{\textbf{long-}\\ \textbf{form}} & \multirowcell{\textbf{short-}\\ \textbf{form}} \\ 
\midrule

\multirow{5}{*}{\multirowcell{Information-\\based}} & Max Claim/Sequence Probability  & \checkmark & \checkmark\\
 & Perplexity \cite{fomicheva-etal-2020-unsupervised} & \checkmark & \checkmark\\
& \multirowcell{\hspace{-1.5em} Mean/Max Token Entropy \\ \hspace{2em}\cite{fomicheva-etal-2020-unsupervised}} & \checkmark & \checkmark \\
& CCP \cite{fadeeva-etal-2024-fact}  & \checkmark & \checkmark \\

\midrule

\multirow{1}{*}{\multirowcell{Reflexive}} & P(True) \cite{kadavath2022language}  & \checkmark & \checkmark\\

\midrule

\multirow{5}{*}{\multirowcell{Sample \\diversity}} & Lexical Similarity~\cite{fomicheva-etal-2020-unsupervised} &  & \checkmark \\
& Degree Matrix~\cite{lin2023generating} &  & \checkmark \\
& Sum of Eigenvalues~\cite{lin2023generating} &  & \checkmark \\
& Semantic Entropy~\cite{kuhn2023semantic} &  & \checkmark \\
& SentenceSAR~\cite{duan-etal-2024-shifting} &  & \checkmark \\



\bottomrule

\end{tabular}
}

\caption{Summary of UQ methods used as baselines.}
\label{tab:lmpoly_methods}

\end{table}

\paragraph{Factuality under faithful condition.}
  When a claim $c$ is assessed as faithful to $\retr$, the LLM may still fail to apply that evidence correctly to the user query. For example, the LLM may simply choose one of the entities mentioned in $\retr$, producing a faithful 
  but
  incorrect answer to the query $\xv$. 

  To account for such errors, in long-form QA, we estimate uncertainty within the faithful branch using a simple \textit{Max Claim Probability} baseline, $p(c \mid \xv, \retr)$. In short-form QA, alternative baselines are more suitable, particularly \textit{Semantic Entropy}~\cite{kuhn2023semantic}, which better captures uncertainty in this scenario (see Section~\ref{sec:ablation_franq_components}).
  
  Therefore, we estimate the factuality for faithful claims with \textit{Max Claim Probability} for long-form QA, and \textit{Semantic Entropy} for short-form QA.

\paragraph{Resulting formula.}
  In summary, we estimate the factuality of the claim $c$ with \franq using the following formula:
  \begin{align}
    \franq(c) = P_\text{faithful}(c, \retr) \, \cdot & \,  \text{UQ}_\text{faith}(c) \label{eq:cuq2}  \\
     + \bigl(1 - P_\text{faithful}(c, \retr)\bigr) \, \cdot & \, \text{UQ}_\text{unfaith}(c),\nonumber
  \end{align}
  where we use \textit{AlignScore} to estimate faithfulness probability $P_\text{faithful}$ and two UQ methods, $\text{UQ}_\text{faith}$ and $\text{UQ}_\text{unfaith}$, selected based on empirical performance for long- and short-form QA scenarios. For long-form QA, we use \textit{Max Claim Probability}~\eqref{eq:claim_prob} for \(\text{UQ}_\text{faith}\) and \textit{Parametric Knowledge}~\eqref{eq:param_know} for \(\text{UQ}_\text{unfaith}\). For short-form QA, we use \textit{Semantic Entropy}~\cite{kuhn2023semantic} for \(\text{UQ}_\text{faith}\) and \textit{Sum of Eigenvalues}~\cite{lin2023generating} for \(\text{UQ}_\text{unfaith}\).

  We consistently apply the same uncertainty methods across all datasets within each QA setting (short- and long-form), and select uncertainty techniques only based on the nature of the task (using token-level likelihoods for long-form QA and sampling-based metrics for short-form QA).

\subsection{Calibrating \franq}
\label{sec:franq_description}
  Since the UQ methods $\text{UQ}_{\text{faith}}$ and $\text{UQ}_{\text{unfaith}}$ of equation~\eqref{eq:cuq2} may have different distributions, to avoid inconsistencies and miscalibration among various UQ measures, we calibrate their outputs using isotonic regression on the training data~\cite{vashurin2025benchmarkinguncertaintyquantificationmethods}. 

  Formally, given training dataset $\DC = \{(u_i, \text{fact}_i)\}_{i=1}^N$ comprising pairs of UQ scores $u_i$ and corresponding binary factuality labels $\text{fact}_i$ for the $N$ claims, we calibrate the UQ scores by fitting a non-decreasing function $f\colon \RR\rightarrow[0,1]$ through isotonic regression, minimizing the squared error:
  \begin{equation}
    \widehat{f} = \arg\min_{f \in \FC} \sum\nolimits_{i=1}^N \bigl(f(u_i) - \text{fact}_i\bigr)^2,
    \label{eq:calibration}
  \end{equation}
  where $\FC$ denotes the set of all non-decreasing functions mapping real numbers to probabilities in the interval $[0,1]$. 
  
  Isotonic regression directly optimizes over $\FC$ without assuming any parametric or functional form for $f$. This yields a piecewise-constant function $\widehat{f}$ defined on the observed UQ scores that satisfies the monotonicity constraint.
  During inference, we apply the calibration function $\widehat{f}$ to each UQ score to produce probabilistically meaningful output.

  \noindent{\bf Condition-Calibrated \franq.} Since $\text{UQ}_\text{faith}$ and $\text{UQ}_\text{unfaith}$ represent factuality scores under faithful and unfaithful conditions, respectively, we propose 
  condition-specific calibration. This involves partitioning the training dataset $\DC$ into two subsets: faithful claims $\DC_\text{faith}$ and unfaithful claims $\DC_\text{unfaith}$. Then, we calibrate $\text{UQ}_\text{faith}$ using the subset $\DC_\text{faith}$ and $\text{UQ}_\text{unfaith}$ using the subset $\DC_\text{unfaith}$.

  We consider \franq with condition-specific calibration as our primary method. To evaluate the impact of calibration, we additionally assess two variants: one without any calibration, and another one in which both UQ methods are calibrated using the full training dataset $\DC$. 
  The calibration strategies are summarized as follows:
  \begin{enumerate}
    \item \textbf{No calibration.} Raw outputs from $\text{UQ}_{\text{faith}}$ and $\text{UQ}_{\text{unfaith}}$ are directly used in equation~\eqref{eq:cuq2} without any calibration.

    \item \textbf{Calibrated.} Both UQ methods are calibrated on the entire training dataset $\DC$, disregarding claim faithfulness.

    \item \textbf{Condition-calibrated.} Each UQ method is calibrated using a subset of the training data corresponding to the respective condition: $\text{UQ}_{\text{faith}}$ is calibrated using $\DC_\text{faith}$, and $\text{UQ}_{\text{unfaith}}$ is calibrated using $\DC_\text{unfaith}$.
  \end{enumerate}


\section{Datasets for RAG Uncertainty Quantification}
\label{sec:rag_datasets}
  Existing datasets for studying RAG hallucinations have serious limitations, as they typically evaluate only context-relative correctness rather than factuality (see Section~\ref{rw}).
  We argue that factuality is more critical in RAG applications, with faithfulness serving as a complementary perspective. Consequently, an effective dataset should capture both factual errors and contextual misuse.
  
  To address this need, we introduce a new dataset specifically designed for long-form generations, enabling fine-grained analysis of atomic claims.
  

\subsection{Long-Form QA Dataset}
  
\paragraph{Questions.} 
  Our long-form QA dataset consists of 76 questions: 44 most challenging questions from RAGTruth~\cite{wu2023ragtruth} (identified by those with highest number of hallucinated claims), and 32 additional technical ``how-to'' questions generated using \gpt via simple prompts (e.g., requesting challenging, domain-diverse technical questions such as \textit{``How does solar power generate electricity?''}). The generated questions were manually inspected to ensure clarity and relevance.

\paragraph{Retrieval Model.}
  For each question, we retrieve the top-$k$=3 passages using the Facebook Contriever~\cite{izacard2021contriever}
  with embeddings from the 2018 English Wikipedia, ensuring high-quality and reliable evidence passages.

\paragraph{LLMs.}
  We construct four model-specific dataset subsets by generating long-form answers to all 76 questions with their corresponding retrieved passages, using greedy decoding independently for each model: \llama, \llamalarge~\cite{grattafiori2024llama3herdmodels}, \falcon~\cite{almazrouei2023falconseriesopenlanguage}, and \gemmasmall~\cite{team2025gemma}. These subsets enable UQ methods to be evaluated on top of in-policy generations for each model.

  \begin{table*}[!ht]
\small
\centering
\resizebox{.95\textwidth}{!}{
\begin{tabular}{l|cc|cc|cc|cc}
 \toprule
  \multirow{2}{*}{\textbf{Method}} & \multicolumn{2}{c|}{\textbf{\llama}} & \multicolumn{2}{c|}{\textbf{\falcon}} & \multicolumn{2}{c|}{\textbf{\llamalarge}} & \multicolumn{2}{c}{\textbf{\gemmasmall}}\\
 & \textbf{PR-AUC $\uparrow$} & \textbf{PRR $\uparrow$} & \textbf{PR-AUC $\uparrow$} & \textbf{PRR $\uparrow$} & \textbf{PR-AUC $\uparrow$} & \textbf{PRR $\uparrow$} & \textbf{PR-AUC $\uparrow$} & \textbf{PRR $\uparrow$} \\
\midrule
 \rowcolor{gray!20}
 \multicolumn{9}{c}{\textit{General Baselines}} \\
\midrule
Max Claim Prob. & .058 {\tiny{$\pm$ .008}} & -.029 {\tiny{$\pm$ .017}} & .126 {\tiny{$\pm$ .007}} & .258 {\tiny{$\pm$ .015}} & .055 {\tiny{$\pm$ .004}} & .118 {\tiny{$\pm$ .023}} & .061 {\tiny{$\pm$ .005}} & .000 {\tiny{$\pm$ .022}} \\
P(True) & .117 {\tiny{$\pm$ .012}} & .207 {\tiny{$\pm$ .023}} & .077 {\tiny{$\pm$ .003}} & .170 {\tiny{$\pm$ .013}} & .071 {\tiny{$\pm$ .009}} & .112 {\tiny{$\pm$ .026}} & .096 {\tiny{$\pm$ .012}} & .148 {\tiny{$\pm$ .018}} \\
Perplexity & .056 {\tiny{$\pm$ .004}} & -.081 {\tiny{$\pm$ .018}} & .090 {\tiny{$\pm$ .004}} & .165 {\tiny{$\pm$ .016}} & .075 {\tiny{$\pm$ .009}} & .090 {\tiny{$\pm$ .024}} & .048 {\tiny{$\pm$ .003}} & -.071 {\tiny{$\pm$ .019}} \\
Max Token Entropy & .109 {\tiny{$\pm$ .004}} & .115 {\tiny{$\pm$ .020}} & .130 {\tiny{$\pm$ .008}} & .219 {\tiny{$\pm$ .016}} & \textbf{.102} {\tiny{$\pm$ .010}} & .138 {\tiny{$\pm$ .023}} & .051 {\tiny{$\pm$ .003}} & -.003 {\tiny{$\pm$ .022}} \\
CCP & .085 {\tiny{$\pm$ .006}} & .169 {\tiny{$\pm$ .024}} & \underline{.162} {\tiny{$\pm$ .010}} & .181 {\tiny{$\pm$ .017}} & .061 {\tiny{$\pm$ .005}} & .108 {\tiny{$\pm$ .022}} & .087 {\tiny{$\pm$ .008}} & \underline{.216} {\tiny{$\pm$ .026}} \\
\midrule
 \rowcolor{gray!20}
 \multicolumn{9}{c}{\textit{RAG-Specific Baselines}} \\
\midrule
AlignScore & .075 {\tiny{$\pm$ .004}} & .108 {\tiny{$\pm$ .020}} & .104 {\tiny{$\pm$ .005}} & .233 {\tiny{$\pm$ .016}} & .068 {\tiny{$\pm$ .007}} & .119 {\tiny{$\pm$ .025}} & .061 {\tiny{$\pm$ .004}} & .058 {\tiny{$\pm$ .021}} \\
Parametric Knowledge & .064 {\tiny{$\pm$ .006}} & .018 {\tiny{$\pm$ .021}} & .067 {\tiny{$\pm$ .003}} & .029 {\tiny{$\pm$ .015}} & .059 {\tiny{$\pm$ .005}} & .047 {\tiny{$\pm$ .023}} & \underline{.112} {\tiny{$\pm$ .011}} & .183 {\tiny{$\pm$ .025}} \\
\midrule
 \rowcolor{gray!20}
 \multicolumn{9}{c}{\textit{XGBoost}} \\
\midrule
XGBoost (all UQ features) & \underline{.124} {\tiny{$\pm$ .006}} & .206 {\tiny{$\pm$ .022}} & .088 {\tiny{$\pm$ .004}} & .198 {\tiny{$\pm$ .014}} & .044 {\tiny{$\pm$ .003}} & -.015 {\tiny{$\pm$ .024}} & .073 {\tiny{$\pm$ .008}} & .085 {\tiny{$\pm$ .023}} \\
XGBoost (\franq features) & .111 {\tiny{$\pm$ .010}} & .149 {\tiny{$\pm$ .020}} & .080 {\tiny{$\pm$ .005}} & .086 {\tiny{$\pm$ .016}} & .048 {\tiny{$\pm$ .003}} & .017 {\tiny{$\pm$ .023}} & .090 {\tiny{$\pm$ .004}} & .158 {\tiny{$\pm$ .022}} \\
\midrule
 \rowcolor{gray!20}
 \multicolumn{9}{c}{\franq} \\
\midrule
\franq no calibration & .100 {\tiny{$\pm$ .007}} & .181 {\tiny{$\pm$ .024}} & .135 {\tiny{$\pm$ .007}} & \textbf{.362} {\tiny{$\pm$ .017}} & .063 {\tiny{$\pm$ .005}} & \underline{.162} {\tiny{$\pm$ .026}} & .080 {\tiny{$\pm$ .005}} & .200 {\tiny{$\pm$ .021}} \\
\franq calibrated & .103 {\tiny{$\pm$ .008}} & \textbf{.256} {\tiny{$\pm$ .020}} & .074 {\tiny{$\pm$ .003}} & .090 {\tiny{$\pm$ .014}} & .043 {\tiny{$\pm$ .003}} & -.047 {\tiny{$\pm$ .022}} & \textbf{.150} {\tiny{$\pm$ .005}} & \textbf{.401} {\tiny{$\pm$ .022}} \\
\franq condition-calibrated & \textbf{.140} {\tiny{$\pm$ .012}} & \underline{.223} {\tiny{$\pm$ .025}} & \textbf{.173} {\tiny{$\pm$ .010}} & \underline{.354} {\tiny{$\pm$ .017}} & \underline{.081} {\tiny{$\pm$ .008}} & \textbf{.184} {\tiny{$\pm$ .027}} & .090 {\tiny{$\pm$ .008}} & .208 {\tiny{$\pm$ .025}} \\
\bottomrule
\end{tabular}
}
\normalsize
\caption{Results on long-form QA benchmark with factuality target. Higher values indicate better performance.
In every setting, the top-performing method is one of the \franq variants.
}
\label{tab:claim_factuality}
\end{table*}
  
  \begin{table*}[!ht]
\small
\centering
\resizebox{.98\textwidth}{!}{
\begin{tabular}{l|cc|cc|cc|cc}
\toprule
\multirow{2}{*}{\textbf{Method}} 
& \multicolumn{2}{c|}{\textbf{\llama}} 
& \multicolumn{2}{c|}{\textbf{\falcon}} 
& \multicolumn{2}{c|}{\textbf{\llamalarge}} 
& \multicolumn{2}{c}{\textbf{\gemma}} \\
 & \textbf{PR-AUC $\uparrow$} & \textbf{PRR $\uparrow$} & \textbf{PR-AUC $\uparrow$} & \textbf{PRR $\uparrow$} & \textbf{PR-AUC $\uparrow$} & \textbf{PRR $\uparrow$} & \textbf{PR-AUC $\uparrow$} & \textbf{PRR $\uparrow$} \\
\midrule
 \rowcolor{gray!20}
 \multicolumn{9}{c}{General Baselines} \\
\midrule
Max Sequence Prob. & .558 {\tiny{$\pm$ .007}} & .454 {\tiny{$\pm$ .008}} & .628 {\tiny{$\pm$ .004}} & .256 {\tiny{$\pm$ .007}} & .569 {\tiny{$\pm$ .007}} & .407 {\tiny{$\pm$ .007}} & .400 {\tiny{$\pm$ .006}} & .162 {\tiny{$\pm$ .009}} \\
Mean Token Entropy & .594 {\tiny{$\pm$ .007}} & .481 {\tiny{$\pm$ .008}} & .613 {\tiny{$\pm$ .004}} & .242 {\tiny{$\pm$ .007}} & .640 {\tiny{$\pm$ .007}} & .491 {\tiny{$\pm$ .008}} & .423 {\tiny{$\pm$ .006}} & .230 {\tiny{$\pm$ .008}} \\
CCP & .551 {\tiny{$\pm$ .007}} & .443 {\tiny{$\pm$ .008}} & .641 {\tiny{$\pm$ .004}} & .304 {\tiny{$\pm$ .007}} & .553 {\tiny{$\pm$ .007}} & .417 {\tiny{$\pm$ .008}} & .412 {\tiny{$\pm$ .006}} & .198 {\tiny{$\pm$ .009}} \\
Lexical Similarity & .564 {\tiny{$\pm$ .008}} & .479 {\tiny{$\pm$ .008}} & .618 {\tiny{$\pm$ .004}} & .277 {\tiny{$\pm$ .007}} & .639 {\tiny{$\pm$ .007}} & .532 {\tiny{$\pm$ .007}} & .430 {\tiny{$\pm$ .007}} & .240 {\tiny{$\pm$ .009}} \\
Degree Matrix & \underline{.629} {\tiny{$\pm$ .008}} & .520 {\tiny{$\pm$ .008}} & .702 {\tiny{$\pm$ .003}} & \underline{.464} {\tiny{$\pm$ .006}} & .627 {\tiny{$\pm$ .007}} & .492 {\tiny{$\pm$ .007}} & .464 {\tiny{$\pm$ .007}} & .260 {\tiny{$\pm$ .009}} \\
Sum of Eigenvalues & .628 {\tiny{$\pm$ .008}} & .518 {\tiny{$\pm$ .008}} & .700 {\tiny{$\pm$ .003}} & .460 {\tiny{$\pm$ .006}} & .628 {\tiny{$\pm$ .007}} & .489 {\tiny{$\pm$ .007}} & .467 {\tiny{$\pm$ .007}} & .260 {\tiny{$\pm$ .009}} \\
Semantic Entropy & .613 {\tiny{$\pm$ .007}} & .525 {\tiny{$\pm$ .008}} & .623 {\tiny{$\pm$ .004}} & .278 {\tiny{$\pm$ .006}} & .637 {\tiny{$\pm$ .007}} & .519 {\tiny{$\pm$ .007}} & .466 {\tiny{$\pm$ .007}} & .261 {\tiny{$\pm$ .008}} \\
SentenceSAR & .571 {\tiny{$\pm$ .007}} & .483 {\tiny{$\pm$ .008}} & .602 {\tiny{$\pm$ .004}} & .263 {\tiny{$\pm$ .006}} & .556 {\tiny{$\pm$ .007}} & .414 {\tiny{$\pm$ .007}} & .416 {\tiny{$\pm$ .006}} & .174 {\tiny{$\pm$ .008}} \\
\midrule
 \rowcolor{gray!20}
 \multicolumn{9}{c}{RAG-specific Baselines} \\
\midrule
AlignScore & .415 {\tiny{$\pm$ .007}} & .207 {\tiny{$\pm$ .009}} & .666 {\tiny{$\pm$ .005}} & .372 {\tiny{$\pm$ .008}} & .432 {\tiny{$\pm$ .007}} & .224 {\tiny{$\pm$ .008}} & .376 {\tiny{$\pm$ .006}} & .158 {\tiny{$\pm$ .008}} \\
Parametric Knowledge & .425 {\tiny{$\pm$ .007}} & .247 {\tiny{$\pm$ .009}} & .556 {\tiny{$\pm$ .005}} & .104 {\tiny{$\pm$ .009}} & .499 {\tiny{$\pm$ .007}} & .330 {\tiny{$\pm$ .008}} & .364 {\tiny{$\pm$ .006}} & .105 {\tiny{$\pm$ .009}} \\
\midrule
 \rowcolor{gray!20}
 \multicolumn{9}{c}{XGBoost} \\
\midrule
XGBoost (all UQ features) & .594 {\tiny{$\pm$ .008}} & .494 {\tiny{$\pm$ .008}} & \underline{.705} {\tiny{$\pm$ .004}} & .462 {\tiny{$\pm$ .007}} & .634 {\tiny{$\pm$ .007}} & .503 {\tiny{$\pm$ .008}} & \underline{.474} {\tiny{$\pm$ .007}} & \textbf{.301} {\tiny{$\pm$ .008}} \\
XGBoost (\franq features) & .526 {\tiny{$\pm$ .008}} & .409 {\tiny{$\pm$ .008}} & .670 {\tiny{$\pm$ .004}} & .368 {\tiny{$\pm$ .007}} & .524 {\tiny{$\pm$ .007}} & .385 {\tiny{$\pm$ .008}} & .414 {\tiny{$\pm$ .006}} & .196 {\tiny{$\pm$ .009}} \\
\midrule
 \rowcolor{gray!20}
 \multicolumn{9}{c}{\franq} \\
\midrule
\franq no calibration & .553 {\tiny{$\pm$ .007}} & .403 {\tiny{$\pm$ .008}} & .641 {\tiny{$\pm$ .003}} & .345 {\tiny{$\pm$ .006}} & .523 {\tiny{$\pm$ .007}} & .340 {\tiny{$\pm$ .008}} & .447 {\tiny{$\pm$ .007}} & .225 {\tiny{$\pm$ .008}} \\
\franq calibrated & .628 {\tiny{$\pm$ .007}} & \underline{.537} {\tiny{$\pm$ .008}} & .672 {\tiny{$\pm$ .003}} & .411 {\tiny{$\pm$ .006}} & \underline{.644} {\tiny{$\pm$ .007}} & \underline{.534} {\tiny{$\pm$ .007}} & .481 {\tiny{$\pm$ .007}} & .258 {\tiny{$\pm$ .009}} \\
\franq condition-calibrated & \textbf{.631} {\tiny{$\pm$ .007}} & \textbf{.541} {\tiny{$\pm$ .008}} & \textbf{.711} {\tiny{$\pm$ .003}} & \textbf{.477} {\tiny{$\pm$ .006}} & \textbf{.647} {\tiny{$\pm$ .007}} & \textbf{.540} {\tiny{$\pm$ .007}} & \textbf{.496} {\tiny{$\pm$ .007}} & \underline{.283} {\tiny{$\pm$ .009}} \\
 \bottomrule
\end{tabular}
}
\caption{Results in PR-AUC$\uparrow$ and PRR$\uparrow$, averaged across four QA datasets for \llama, \falcon, \llamalarge and \gemma.
The condition-calibrated \franq is top-performing across all settings, except mean PRR on \gemma, where it ranks second.}
\label{tab:qa_meanrank}
\end{table*}
  
  \begin{figure*}
    \centering
    \begin{subfigure}{0.49\linewidth}
      \centering
      \includegraphics[width=\linewidth]{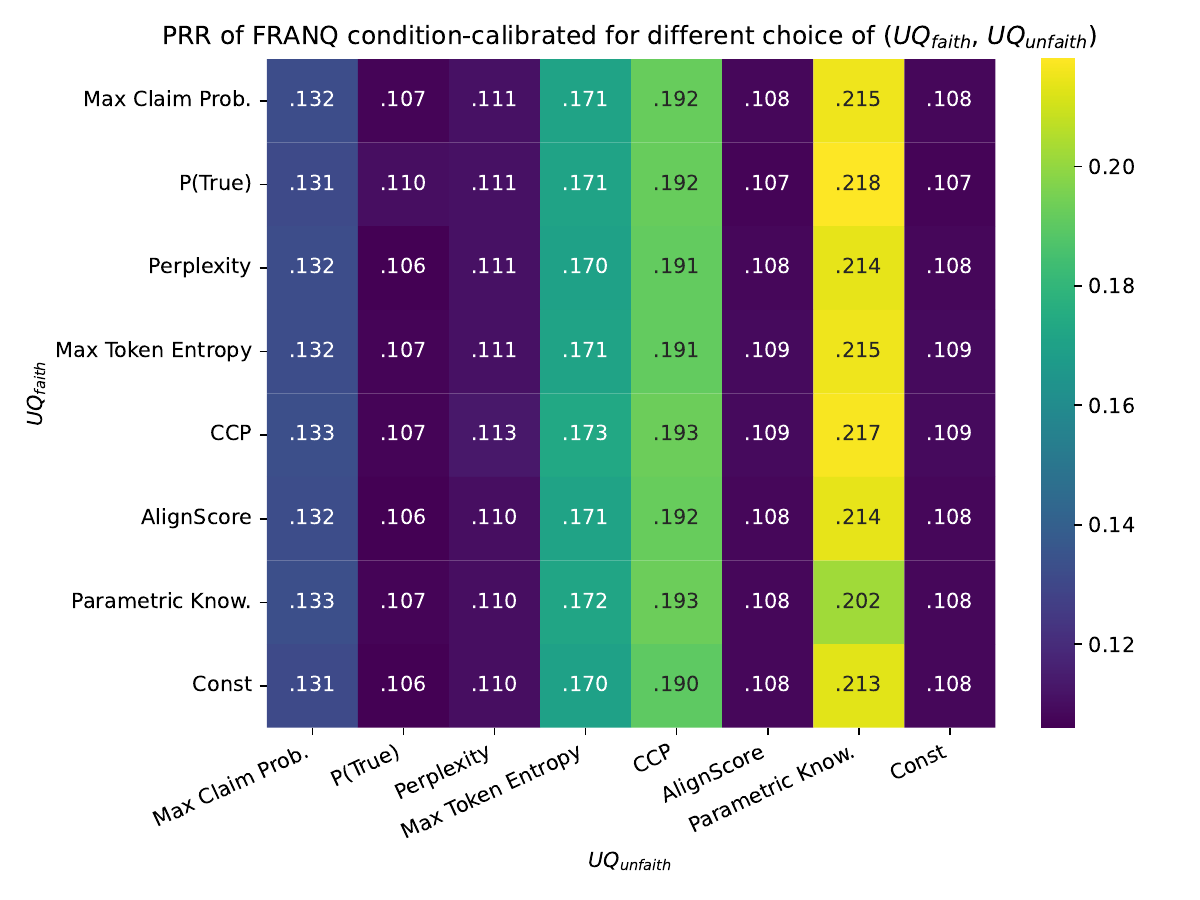}
      \caption{PRR on long-form QA dataset.}
    \end{subfigure}
    \hfill
    \begin{subfigure}{0.49\linewidth}
      \centering
      \includegraphics[width=\linewidth]{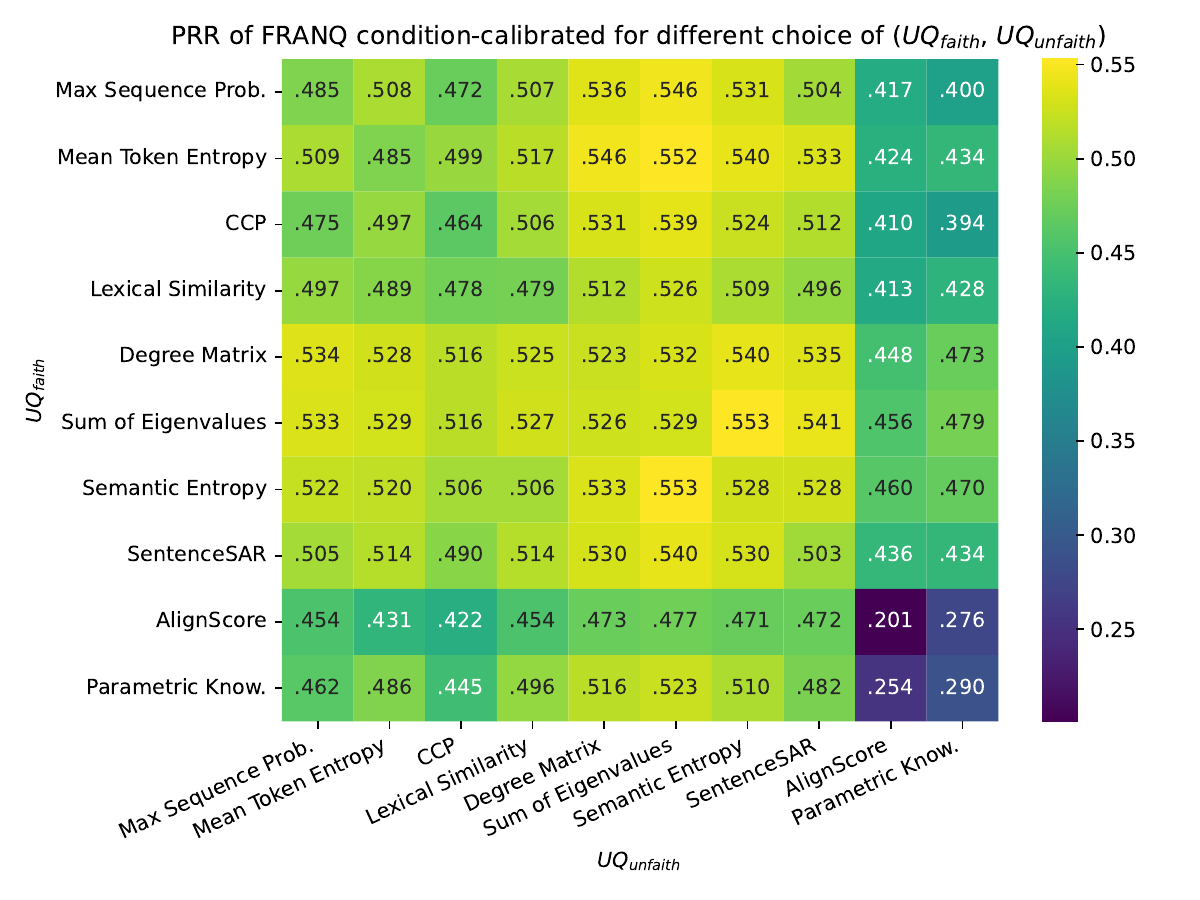}
      \caption{PRR on short-form QA benchmark (mean across 4 datasets).}
    \end{subfigure}
    \caption{PRR of condition-calibrated \franq for different choices of $\text{UQ}_\text{faith}$ and $\text{UQ}_\text{unfaith}$.}
    \label{fig:ablations_uqfaith_uqunfaith}
  \end{figure*}

\paragraph{Claim Extraction.}
  For each generated answer, we extract atomic claims and their corresponding token spans using the approach of~\cite{wang-etal-2024-factcheck, vashurin2025benchmarkinguncertaintyquantificationmethods}. Following prior work, \gpto first extracts decontextualized atomic claims from the full paragraph through a dedicated prompt. Then, for each claim, a second prompt identifies the corresponding words in the original text, which we map to token spans. Applying this procedure, we obtain 1,782 claims for \llama and 1,548 claims for \falcon. From these claims, we select 500 claims for train set, reserving the remainder for test set. The prompts used for claim extraction and mapping are listed in Appendix~\ref{sec:generation_prompts_and_setup}.

\paragraph{Annotation.}
  We annotate each atomic claim for both faithfulness and factuality using a two-stage procedure. In the first stage, we use \gpto-search, a \gpto-based search model augmented with web search over up-to-date web sources, to assign faithfulness labels (\textit{faithful} or \textit{unfaithful}) and factuality labels (\textit{True}, \textit{False}, or \textit{Unverifiable}) using dedicated prompts. 
  
  In the second stage, we manually review all claims initially labeled as \textit{False} or \textit{Unverifiable}, which are the most difficult cases for automatic annotation, and correct the labels when necessary. This targeted verification step helps improve the overall quality and reliability of the annotations. We then retain only verifiable claims (\textit{True} or \textit{False}) and binarize the labels accordingly.

  This design is supported by our validation analysis: we found that the automatic annotation is substantially more reliable for \textit{True} claims than for \textit{False}/\textit{Unverifiable} ones (90\% vs.\ $\sim$40\% precision), which motivates targeted manual verification of the latter. On the \llama subset, this second stage improves overall annotation accuracy from 78\% to 91\%. Human inter-annotator accuracy for factuality is 0.87, indicating strong agreement among the annotators. Further details on the prompts, annotation protocol, dataset statistics, and agreement/error analysis are provided in Appendix~\ref{sec:generation_prompts_and_setup}.


  
  
\subsection{Short-form QA Datasets}
  In contrast to long-form QA, where evaluating factuality requires extracting model-specific claims and annotating them, short-form QA provides gold-standard answers for each question. This allows us to directly compare each model's generated answer with the ground-truth answer, yielding an automatic factuality judgment without additional manual annotation or claim-level verification.

\paragraph{Questions.}
  We adapt four short-form QA datasets for RAG evaluation: TriviaQA~\cite{joshi-etal-2017-triviaqa}, SimpleQA~\cite{wei2024measuringshortformfactualitylarge}, Natural Questions~\cite{kwiatkowski-etal-2019-natural}, and PopQA~\cite{mallen2022not}. For each dataset, we sample 200 questions for training and 1000 for testing, and we treat each model response as a single claim.

\paragraph{RAG Models.}
  We use the same retrieval model as in the long-form setting, selecting the top-k=5 passages per question. For LLMs, we use the same two Llama models and the Falcon model, along with an additional model: \gemma~\cite{team2025gemma}.

\paragraph{Annotation.}
  We evaluate factuality of each generated answer by comparing it against the gold-standard answer using \gpto{}, following the procedure of~\cite{wei2024measuringshortformfactualitylarge}, which has been shown to yield reliable factuality judgments.

\section{Experiments}
\label{sec:experiments}
  In this section, we evaluate \franq and corresponding baselines on both the short-form and long-form benchmarks described in Section~\ref{sec:rag_datasets}. For all experiments, we fix the retrieval process and the underlying white-box LLM, and we assess the factual accuracy of the model-generated claims. This allows us to isolate the effect of different uncertainty estimation approaches.

  Later, through ablation studies, we examine the contribution of the individual \franq components, $P(\text{faithful})$, $\text{UQ}_{\text{faith}}$, and $\text{UQ}_{\text{unfaith}}$, as well as the effect of varying the amount of training data.

\subsection{Experimental Setup}

\noindent{\bf UQ baselines.}
  We group all UQ methods into four categories: (1)~general baselines, (2)~RAG-specific baselines, (3)~XGBoost-based methods, and (4)~three variants of our proposed \franq method, each using a different calibration strategy.

  \textit{General baselines.} We compare \franq with general baselines, which consist of standard UQ methods applied directly to the LLM's output distribution without using any RAG-specific structure. For implementation, we use the LM-Polygraph library~\cite{fadeeva2023lmpolygraphuncertaintyestimationlanguage}. A complete list of methods we used is provided in Table~\ref{tab:lmpoly_methods}.
  
  \textit{RAG-specific baselines.} We also evaluate the two \franq components in isolation, \textit{AlignScore} and \textit{Parametric Knowledge}, to assess how much their combination in \franq improves over using each component individually (see Section~\ref{sec:franq_components}).

  \textit{XGBoost methods.} We include XGBoost models trained on factuality labels using two feature sets: (1)~the three components used in \franq (\textit{AlignScore}, $\text{UQ}_\text{faith}$, $\text{UQ}_\text{unfaith}$), and (2)~all available unsupervised UQ method.

  \textit{FRANQ.} Finally, we evaluate three \franq variants with different calibration strategies for $\text{UQ}_\text{faith}$ and $\text{UQ}_\text{unfaith}$ (see Section~\ref{sec:franq_description}): \textit{no calibration}, \textit{calibrated}, and \textit{condition-calibrated}.

\paragraph{Evaluation measures.}
  Each UQ method produces factuality estimates, which we compare against binary gold-standard labels using PR-AUC, treating false claims as the positive class to emphasize their detection. We also assess rejection performance using the Prediction Rejection Ratio (PRR; \citealp{mallen2022not}) with a maximum rejection threshold of 0.5. PRR measures how effectively the model rejects uncertain predictions while retaining accurate ones, capturing its ability to prioritize reliable outputs.

\subsection{Long-Form QA Results}
  For long-form QA, we evaluate each UQ method using PR-AUC and PRR across four models (\llama, \falcon, \llamalarge and \gemmasmall), see Table~\ref{tab:claim_factuality}.
  The condition-calibrated \franq achieves the best PR-AUC and second-best PRR for \llama and \falcon, while for \llamalarge it attains the best PRR and second-best PR-AUC. 
  
  The calibrated \franq achieves the highest PRR for \llama and the highest PR-AUC and PRR for \gemmasmall. The non-calibrated \franq also performs strongly, ranking first and second in PRR for \falcon and \llamalarge, respectively.
  Overall, \franq demonstrates strong and consistent performance across all models.

\subsection{Short-Form QA Results}
  For short-form QA, we evaluate UQ methods using PR-AUC and PRR across four models (\llama, \llamalarge, \falcon, \gemma) and four datasets. To account for dataset variability, we report mean scores averaged over datasets, following~\citet{vashurin2025benchmarkinguncertaintyquantificationmethods} (Table~\ref{tab:qa_meanrank}); per-dataset results appear in Appendix~\ref{sec:qa_results_all}.

  Condition-calibrated \franq achieves the best mean performance across all models and both measures, except for mean PRR on \gemma, where it ranks second.
  Calibrated \franq ranks second for PRR on \llama and for both PR-AUC and PRR on \llamalarge.
  Among unsupervised methods, Degree Matrix and Lexical Similarity perform strongly, ranking second-best in several settings.

  \begin{table}[t!]
\centering
\resizebox{1.0\linewidth}{!}{
\begin{tabular}{l|cc|cc}
\toprule
\multirow{2}{*}{\textbf{Method}} 
& \multicolumn{2}{c|}{\textbf{Shuffled retrievals}}
& \multicolumn{2}{c}{\textbf{Corrupted retrievals}} \\
& \textbf{PR-AUC $\uparrow$} & \textbf{PRR $\uparrow$} & \textbf{PR-AUC $\uparrow$} & \textbf{PRR $\uparrow$} \\
\midrule
\rowcolor{gray!20}
\multicolumn{5}{c}{General Baselines} \\
\midrule
Max Sequence Prob. & .638 & .489 & .776 & .374 \\
Mean Token Entropy & .651 & .460 & .767 & .358 \\
CCP & .629 & .472 & .779 & .392 \\
Lexical Similarity & .668 & .512 & .767 & .358 \\
Degree Matrix & .687 & .548 & .781 & .406 \\
Sum of Eigenvalues & .686 & .536 & .782 & .404 \\
Semantic Entropy & .666 & .537 & .780 & .447 \\
SentenceSAR & .648 & .523 & .772 & .381 \\
\midrule
\rowcolor{gray!20}
\multicolumn{5}{c}{RAG-specific Baselines} \\
\midrule
AlignScore & .507 & .234 & .659 & -.028 \\
Parametric Knowledge & .502 & .220 & .801 & .412 \\
\midrule
\rowcolor{gray!20}
\multicolumn{5}{c}{XGBoost} \\
\midrule
XGBoost (all UQ features) & .684 & .544 & .774 & .354 \\
XGBoost (FRANQ features) & .579 & .402 & \textbf{.813} & .471 \\
\midrule
\rowcolor{gray!20}
\multicolumn{5}{c}{\franq} \\
\midrule
FRANQ no calibration & .586 & .393 & .768 & .331 \\
FRANQ calibrated & \underline{.692} & \underline{.549} & \underline{.807} & \underline{.475} \\
FRANQ condition-calibrated & \textbf{.695} & \textbf{.553} & \textbf{.813} & \textbf{.478} \\
\bottomrule
\end{tabular}
}
\caption{Robustness of \llama under two retrieval corruption settings on four short-form QA datasets. In \textit{shuffled retrievals}, 50\% of passages are replaced with unrelated ones; in \textit{factually corrupted retrievals}, 50\% are modified to contain incorrect facts.}
\vspace{-1.0em}
\label{tab:retrieval_corruptions}
\end{table}

\subsection{Ablation Studies}
\label{sec:ablation_franq_components}
  In this section, we summarize the main observations from ablation studies examining (1) the contribution of \franq's components, (2) robustness to retrieval noise, (3) the effect of supervision and (4) computational efficiency. Complete experimental descriptions, tables, and additional ablations are provided in Appendix~\ref{sec:appendix_ablations}.

\noindent{\bf Analysis of \franq's components.}
  Figure~\ref{fig:ablations_uqfaith_uqunfaith} shows the PRR of condition-calibrated \franq for different choices of $\text{UQ}_\text{faith}$ and $\text{UQ}_\text{unfaith}$, evaluated with \llama on long-form QA and on a subset of 200 questions from each short-form dataset.

  On long-form QA (Figure~\ref{fig:ablations_uqfaith_uqunfaith}(a)), performance is largely insensitive to the choice of $\text{UQ}_\text{faith}$, whereas the choice of $\text{UQ}_\text{unfaith}$ is critical: using Parametric Knowledge as $\text{UQ}_\text{unfaith}$ yields the strongest PRR for nearly all $\text{UQ}_\text{faith}$ options. This suggests that, in long-form QA, modeling factuality for unfaithful claims is the key design choice.

  On short-form QA (Figure~\ref{fig:ablations_uqfaith_uqunfaith}(b)), many combinations perform similarly, indicating that \franq is relatively robust to these choices. The configuration used in our short-form experiments, Semantic Entropy for $\text{UQ}_\text{faith}$ and Sum of Eigenvalues for $\text{UQ}_\text{unfaith}$, achieves the best observed PRR of 0.553.

  Additional faithfulness ablations are reported in Appendix~\ref{sec:ablations_thresholded_faith}. In particular, replacing continuous AlignScore probabilities with binary thresholding consistently degrades performance, highlighting the value of probabilistic faithfulness weighting.
  
  \begin{table}[t!] 
\centering
\resizebox{1.0\linewidth}{!}{
\small
\begin{tabular}{lccc}
\toprule
\textbf{Method} & \multirowcell{\bf Inference\\ \bf Runtime} & \multirowcell{\bf Training\\ \bf Time} & \multirowcell{\bf Model\\ \bf Size} \\
\midrule
\rowcolor{gray!20}
\multicolumn{4}{c}{General Baselines} \\
\midrule
Max Claim Probability      & $<0.1$ s & — & — \\
P(True)                    & 1.3 s    & — & — \\
Perplexity                 & $<0.1$ s & — & — \\
Max Token Entropy          & $<0.1$ s & — & — \\
CCP                        & 1.7 s    & — & — \\
\midrule
\rowcolor{gray!20}
\multicolumn{4}{c}{RAG-specific Baselines} \\
\midrule
AlignScore                 & 0.5 s    & — & — \\
Parametric Knowledge       & 1.6 s    & — & — \\
\midrule
\rowcolor{gray!20}
\multicolumn{4}{c}{XGBoost} \\
\midrule
XGBoost (all UQ features)  & 1.9 s    & 0.60 s & 10 kB \\
XGBoost (\franq features)   & 1.7 s    & 0.12 s & 14 kB \\
\midrule
\rowcolor{gray!20}
\multicolumn{4}{c}{\franq} \\
\midrule
\franq (no calibration)       & 1.7 s & —      & — \\
\franq (calibrated)           & 1.7 s & 0.57 s & 312 B \\
\franq (condition-calibrated) & 1.7 s & 0.58 s & 244 B \\
\bottomrule
\end{tabular}
}
\caption{Inference runtime (averaged across dataset samples), training cost, and model size of uncertainty estimators and \franq variants on \llama for short-form QA, measured beyond base LLM generation.}
\vspace{-1.0em}
\label{tab:runtime_analysis}
\end{table}

\paragraph{Robustness to retrieval corruption.}
  We evaluate \franq under two retrieval failure modes. First, we simulate irrelevant retrievals by randomly replacing 50\% of retrieved passages across the four short-form QA datasets, ensuring that no corrupted example retains its original retrieval set. This reduces answer accuracy by 7\% on average.

  Second, we consider a more challenging setting in which retrieved passages remain relevant to the input question but contain subtle factual errors. We construct this setting by using \gpto to rewrite 50\% of retrieved passages with plausible inaccuracies, such as altered dates, people, places, or events. This encourages the LLM to produce answers that are faithful to the retrieval yet factually wrong. In this regime, $\text{UQ}_{\text{faith}}$ must assess whether the retrieval itself is trustworthy. We find that Parametric Knowledge is the most effective choice for $\text{UQ}_{\text{faith}}$, as it captures the model's confidence in the retrieved content.

  Table~\ref{tab:retrieval_corruptions} shows that under both shuffled and factually corrupted retrievals, condition-calibrated and calibrated \franq rank first and second, respectively, across evaluation metrics.  Overall, these results suggest that \franq is robust to both irrelevant and misleading retrieved evidence.

  \begin{figure*}[!ht]
    \centering
    \begin{subfigure}{0.49\linewidth}
        \centering
        \includegraphics[width=\linewidth]{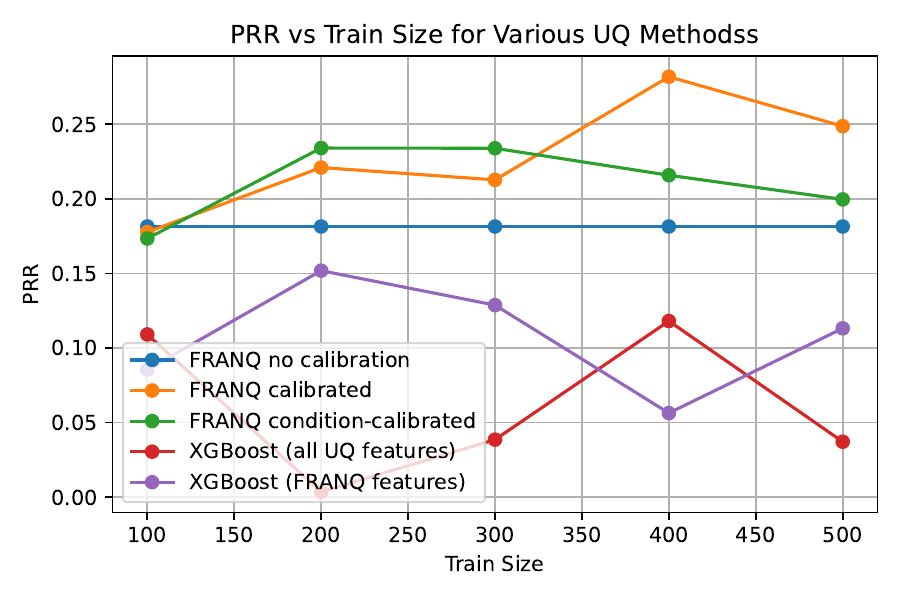}
        \caption{Long-form QA, \llama}
    \end{subfigure}
    \hfill
    \begin{subfigure}{0.49\linewidth}
        \centering
        \includegraphics[width=\linewidth]{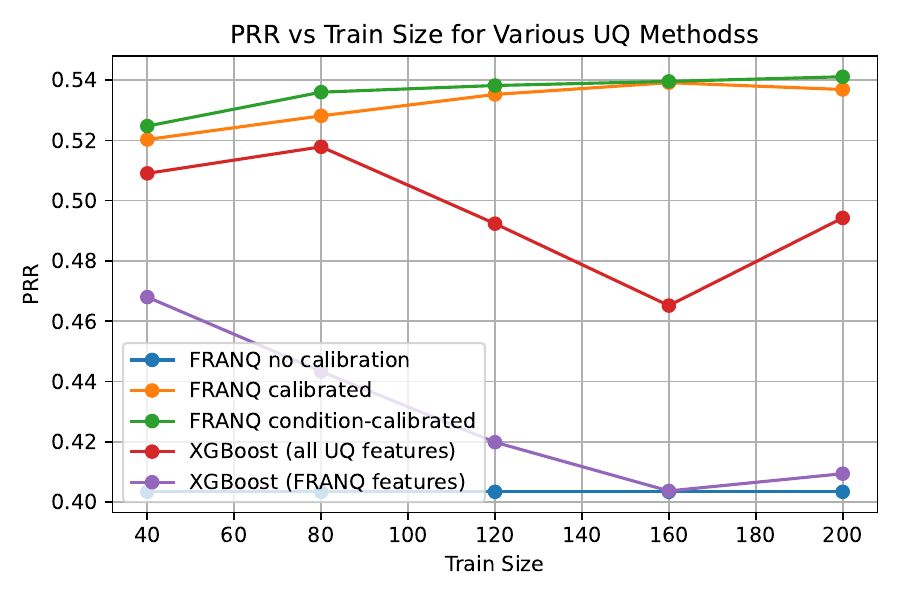}
        \caption{Short-form QA, \llama}
    \end{subfigure}
    \caption{PRR comparison of \franq and XGBoost methods across different training set sizes.}
    \label{fig:ablations_train_size}
  \end{figure*}

\paragraph{Effect of supervision.}
  Figure~\ref{fig:ablations_train_size} shows PRR versus training set size for three \franq variants and two XGBoost baselines on long-form and short-form QA. As expected, uncalibrated \franq is unaffected by training size, while the supervised variants improve with more labeled calibration data and then saturate. On long-form QA, condition-calibrated \franq peaks at around 300 training instances; on short-form QA, performance stabilizes at around 120. Across all training sizes, the calibrated \franq variants consistently outperform the XGBoost baselines.

\paragraph{Computational efficiency.}
  Table~\ref{tab:runtime_analysis} reports the runtime overhead, training cost, and model size of \franq's uncertainty components beyond a completed \llama forward pass on short-form Natural Questions (2.2\,s per instance on average). Overall, \franq adds only modest inference cost. The supervised variants are especially lightweight: isotonic calibration fits in under one second and yields models of only a few hundred bytes, making calibrated \franq practical.

\section{Related Work}\label{rw}

\paragraph{Uncertainty Quantification for RAG.}
  Several UQ methods for RAG study how retrieved knowledge shapes LLM outputs, including lookback-ratio classifiers~\citep{chuang2024lookbacklensdetectingmitigating}, feature-based regression models comparing retrieved and parametric knowledge contributions~\citep{sun2024redeep}, uncertainty estimation based on the signal-to-noise ratio of output probabilities across samples~\citep{li2024uncertaintyrag}, and prompt--response relevance modeling~\citep{hu2024lrp4rag}. 
  These approaches mainly assess hallucinations relative to retrieved context, rather than assessing whether they are factually correct in an absolute sense.

  However, they often incur additional computational cost and do not directly address factual correctness beyond the retrieved evidence. Search-based approaches such as SAFE~\cite{wei2024long} extend verification using LLM agents and web search, but at substantially greater complexity. In contrast, \franq is a lightweight, self-contained UQ framework that combines faithfulness to retrieved context with truthfulness under both faithful and unfaithful conditions, without additional training or external verification.

When retrieval is absent, uncertainty is estimated from internal knowledge using white-box~\citep{fomicheva-etal-2020-unsupervised, kadavath2022language, kuhn2023semantic, fadeeva-etal-2024-fact, duan-etal-2024-shifting} or black-box methods~\citep{fomicheva-etal-2020-unsupervised, lin2023generating}. However, these are studied in isolation and ignore interactions with retrieved evidence. Our work unifies both by jointly modeling retrieval-related and intrinsic uncertainty.

\paragraph{Factuality/Hallucination Datasets for RAG.}
Progress in RAG hallucination detection relies on datasets with reliable factuality annotations. RAGTruth~\citep{wu2023ragtruth} provides span-level labels but excludes factually correct content unsupported by retrieval. Dialogue datasets such as Wizard of Wikipedia~\cite{dinan2018wizard} and FaithDial~\citep{dziri2022faithdial} emphasize grounding, treating unsupported content as hallucinated even if correct. Similarly, QA benchmarks such as RAGBench~\citep{friel2024ragbench} and AdaptiveRAG~\citep{moskvoretskii2025adaptive} define hallucinations relative to context. In contrast, \franq focuses on factuality beyond retrieved evidence.


\section{Conclusion and Future Work}

We introduced \franq, a method for quantifying factuality of claims in RAG outputs via faithfulness. Across long- and short-form QA tasks and multiple LLMs, \franq outperforms unsupervised UQ baselines, RAG-specific methods, and supervised classifiers. We also presented a QA dataset annotated for factuality and faithfulness using a hybrid labeling process. Our approach opens directions for future work, including using \franq's uncertainty signals for generation-time control to enable more reliable RAG systems.


\section*{Limitations}
  While \franq achieves strong hallucination detection performance on average, it does not guarantee perfect factuality estimation in every case, especially in difficult settings involving ambiguous claims, limited evidence, or highly uncertain generations.

  Although \franq is robust to both irrelevant and factually corrupted retrievals, its performance still depends on the quality of the underlying signals used to estimate faithfulness and factuality. In particular, severe retrieval errors or model miscalibration may reduce performance in challenging real-world settings.

  Since some \franq variants rely on calibration of their components, they require a small amount of labeled data. However, our experiments show that these variants remain effective with limited supervision and outperform stronger supervised baselines.


  \section*{Ethical Considerations}
  \franq is designed to reduce the spread of factual errors by improving the reliability and interpretability of language model outputs. By distinguishing between factuality and faithfulness, it can avoid penalizing factually correct but unsupported claims. However, \franq does not prevent hallucinations directly and instead relies on downstream filtering, so its impact depends on how it is integrated into larger systems.

  Although \franq is robust to noisy and factually corrupted retrievals, its performance still depends on retrieval quality. In real-world applications, retrieved documents may be biased, outdated, or incorrect, which can affect the method's output. Careful source selection and monitoring remain important to avoid reinforcing misinformation or harmful biases.

  The long-form evaluation pipeline relies partly on GPT-4o-based claim extraction and annotation. While we combine automatic annotation with targeted manual verification, some biases from the underlying model may still persist. Future work could explore more diverse annotation strategies, including broader human validation.

  Better factuality estimation can support safer deployment of AI systems in knowledge-intensive domains such as education, healthcare, and law. Still, \franq should be viewed as a decision-support tool rather than a replacement for human fact-checking.


  \bibliography{custom}

  \appendix


  \begin{table*}[!t] 
\resizebox{1.0\textwidth}{!}{
\begin{tabular}{l|lll|lll|lll|lll}
\toprule
 \multirow{2}{*}{\textbf{Method}} & \multicolumn{3}{c|}{\textbf{NQ}} & \multicolumn{3}{c|}{\textbf{PopQA}} & \multicolumn{3}{c|}{\textbf{TriviaQA}} & \multicolumn{3}{c}{\textbf{SimpleQA}} \\
 & \textbf{AUROC $\uparrow$} & \textbf{PR-AUC$\uparrow$} & \textbf{PRR$\uparrow$} & \textbf{AUROC $\uparrow$} & \textbf{PR-AUC$\uparrow$} & \textbf{PRR$\uparrow$} & \textbf{AUROC $\uparrow$} & \textbf{PR-AUC$\uparrow$} & \textbf{PRR$\uparrow$} & \textbf{AUROC $\uparrow$} & \textbf{PR-AUC$\uparrow$} & \textbf{PRR$\uparrow$} \\
\midrule
 \rowcolor{gray!20}
 \multicolumn{13}{c}{General Baselines} \\
\midrule
Max Sequence Prob. & .680 & .440 & .292 & .745 & .550 & .421 & .774 & .529 & .478 & .833 & .712 & .625 \\
Mean Token Entropy & .723 & .503 & .389 & \underline{.768} & \textbf{.607} & .455 & .796 & .569 & .523 & .809 & .697 & .555 \\
CCP & .705 & .471 & .357 & .709 & .526 & .393 & .767 & .528 & .471 & .800 & .680 & .552 \\
Lexical Similarity & .720 & .494 & .386 & .763 & .571 & .462 & .775 & .508 & .485 & .818 & .685 & .585 \\
Degree Matrix & \textbf{.751} & \textbf{.557} & \underline{.421} & .738 & .570 & .421 & .816 & \textbf{.626} & .570 & .852 & .764 & .668 \\
Sum of Eigenvalues & .749 & \underline{.553} & .411 & .740 & .564 & .416 & .816 & .621 & .561 & .861 & \underline{.774} & .686 \\
Semantic Entropy & .727 & .518 & .373 & \textbf{.776} & .602 & \textbf{.496} & .801 & .565 & .546 & .863 & .766 & .684 \\
SentenceSAR & .678 & .395 & .269 & .762 & .562 & .459 & .794 & .560 & .521 & .858 & .767 & .682 \\
\midrule
 \rowcolor{gray!20}
 \multicolumn{13}{c}{RAG-specific Baselines} \\
\midrule
AlignScore & .682 & .427 & .312 & .566 & .371 & .079 & .631 & .387 & .215 & .645 & .473 & .221 \\
Parametric Knowledge & .626 & .371 & .203 & .664 & .470 & .290 & .727 & .467 & .397 & .490 & .393 & .096 \\
\midrule
 \rowcolor{gray!20}
 \multicolumn{13}{c}{XGBoost} \\
\midrule
XGBoost (all UQ features) & .712 & .504 & .375 & .744 & .565 & .433 & .773 & .546 & .486 & .835 & .760 & .683 \\
XGBoost (\franq features) & .651 & .412 & .283 & .690 & .503 & .350 & .692 & .441 & .328 & .860 & .747 & .676 \\
\midrule
 \rowcolor{gray!20}
 \multicolumn{13}{c}{\franq} \\
\midrule
\franq no calibration & .637 & .456 & .268 & .676 & .481 & .278 & .773 & .557 & .467 & .826 & .717 & .601 \\
\franq calibrated & .735 & .529 & .405 & .765 & .597 & .468 & \textbf{.821} & \underline{.623} & \textbf{.580} & \underline{.869} & .761 & \underline{.695} \\
\franq condition-calibrated & .748 & .526 & .409 & .763 & \underline{.605} & \underline{.477} & \underline{.821} & .618 & \underline{.576} & \textbf{.877} & \textbf{.776} & \textbf{.703} \\
 \bottomrule
\end{tabular}
}
\caption{Results on 4 QA datasets for \llama.}
\label{tab:qa_all_llama}
\end{table*}
  \begin{table*}[!t]
\resizebox{1.0\textwidth}{!}{
\begin{tabular}{l|lll|lll|lll|lll}
\toprule
 \multirow{2}{*}{\textbf{Method}} & \multicolumn{3}{c|}{\textbf{NQ}} & \multicolumn{3}{c|}{\textbf{PopQA}} & \multicolumn{3}{c|}{\textbf{TriviaQA}} & \multicolumn{3}{c}{\textbf{SimpleQA}} \\
 & \textbf{AUROC $\uparrow$} & \textbf{PR-AUC$\uparrow$} & \textbf{PRR$\uparrow$} & \textbf{AUROC $\uparrow$} & \textbf{PR-AUC$\uparrow$} & \textbf{PRR$\uparrow$} & \textbf{AUROC $\uparrow$} & \textbf{PR-AUC$\uparrow$} & \textbf{PRR$\uparrow$} & \textbf{AUROC $\uparrow$} & \textbf{PR-AUC$\uparrow$} & \textbf{PRR$\uparrow$} \\
\midrule
 \rowcolor{gray!20}
 \multicolumn{13}{c}{General Baselines} \\
\midrule
Max Sequence Prob. & .599 & .555 & .186 & .653 & .649 & .259 & .590 & .487 & .163 & .625 & .820 & .416 \\
Mean Token Entropy & .599 & .542 & .184 & .657 & .662 & .279 & .557 & .432 & .108 & .656 & .814 & .396 \\
CCP & .632 & .576 & .258 & .659 & .648 & .297 & .620 & .518 & .212 & .635 & .822 & .448 \\
Lexical Similarity & .581 & .486 & .115 & .721 & .691 & .412 & .587 & .476 & .157 & .650 & .818 & .422 \\
Degree Matrix & .653 & .571 & .258 & \underline{.787} & \underline{.777} & \textbf{.571} & .660 & .565 & .311 & \textbf{.795} & \textbf{.896} & \textbf{.718} \\
Sum of Eigenvalues & .651 & .568 & .260 & \textbf{.789} & \textbf{.780} & \underline{.570} & .661 & .559 & .299 & \underline{.791} & \underline{.894} & \underline{.713} \\
Semantic Entropy & .561 & .494 & .086 & .718 & .698 & .415 & .584 & .468 & .155 & .685 & .831 & .456 \\
SentenceSAR & .509 & .455 & .012 & .755 & .707 & .463 & .523 & .395 & .026 & .739 & .850 & .552 \\
\midrule
 \rowcolor{gray!20}
 \multicolumn{13}{c}{RAG-specific Baselines} \\
\midrule
AlignScore & .655 & \underline{.613} & .320 & .639 & .652 & .262 & .685 & .540 & \underline{.341} & .748 & .860 & .566 \\
Parametric Knowledge & .556 & .486 & .089 & .611 & .590 & .210 & .567 & .420 & .086 & .512 & .729 & .030 \\
\midrule
 \rowcolor{gray!20}
 \multicolumn{13}{c}{XGBoost} \\
\midrule
XGBoost (all UQ features) & \textbf{.679} & \textbf{.617} & \textbf{.340} & .772 & .748 & .507 & \underline{.693} & \underline{.572} & .340 & .787 & .885 & .661 \\
XGBoost (\franq features) & .640 & .596 & .292 & .694 & .712 & .414 & .624 & .517 & .236 & .731 & .853 & .532 \\
\midrule
 \rowcolor{gray!20}
 \multicolumn{13}{c}{\franq} \\
\midrule
\franq no calibration & .576 & .496 & .113 & .732 & .716 & .448 & .609 & .492 & .205 & .738 & .862 & .616 \\
\franq calibrated & .617 & .541 & .215 & .773 & .749 & .520 & .626 & .513 & .228 & .769 & .885 & .682 \\
\franq condition-calibrated & \underline{.668} & .591 & \underline{.331} & .781 & .764 & .533 & \textbf{.695} & \textbf{.606} & \textbf{.377} & .776 & .886 & .668 \\
 \bottomrule
\end{tabular}
}
\caption{Results on 4 QA datasets for \falcon.}
\label{tab:qa_all_falcon}
\end{table*}

\newpage

\section{Additional Short-Form QA Results}
\label{sec:qa_results_all}


In Table~\ref{tab:qa_meanrank} of the main text, we reported aggregated results for short-form QA using mean values for ease of presentation and to facilitate direct comparison across methods and models in a concise and interpretable manner. Here, we provide the full results for each of the four QA datasets (Natural Questions, PopQA, TriviaQA, SimpleQA) for both \llama (see Table~\ref{tab:qa_all_llama}) and \falcon (see Table~\ref{tab:qa_all_falcon}), offering a more detailed breakdown of performance across datasets and highlighting dataset-specific trends and variations in method behavior, as well as enabling more fine-grained analysis of individual model performance and better understanding of method robustness across different question types.


  For \llama, \franq calibrated and \franq condition-calibrated rank among top performers, including top two on TriviaQA and SimpleQA. On PopQA, \franq condition-calibrated ranks among top three with Semantic Entropy and Max Token Entropy. On Natural Questions, it ranks in top four with DegreeMatrix, Eccentricity, and Sum of Eigenvalues. Overall, both \franq variants achieve best average performance.


For \falcon, \franq condition-calibrated achieves top performance on TriviaQA and second-best on Natural Questions. It ranks among the top three methods on PopQA and top four on SimpleQA, alongside Degree Matrix, Sum of Eigenvalues, and XGBoost (all features). On average, \franq condition-calibrated is the leading method across the datasets.

\newpage
\clearpage

\section{Prompts and Setup}
\label{sec:generation_prompts_and_setup}

\subsection{Short-form QA}
\label{sec:generation_prompts_and_setup_sequence}
  For short-form QA experiments, we paired each question with the top-5 retrieved Wikipedia passages and used the prompt format in Figure~\ref{fig:qa_prompt}. For annotation, \gpto was given the question, model-generated answer, and gold answer, and asked to label responses as correct, incorrect, or not attempted (excluded from evaluation), following~\citet{wei2024measuringshortformfactualitylarge}. Table~\ref{tab:stats_sequence} reports dataset statistics.

\subsection{Long-form QA}
\label{sec:generation_prompts_and_setup_claim}

  For long-form QA experiments, we used each question with the top-3 retrieved Wikipedia passages. All models followed the prompt format shown in Figure~\ref{fig:claim_prompt}. Extracted answers were decomposed into atomic claims using the prompt in Figure~\ref{fig:claim_extraction_prompt}, and each claim was matched to its corresponding span in the original sentence using Figure~\ref{fig:claim_matching_prompt}. Claims without identifiable spans (e.g., due to annotation inconsistencies) were excluded. The remaining claims were annotated for factuality and faithfulness using automatic annotation followed by manual validation (Appendix~\ref{sec:annotation}, \ref{sec:manual_annotation}). Table~\ref{tab:stats_claim} reports dataset statistics.

  Compared to prior decomposition methods such as FActScore~\cite{min-etal-2023-factscore}, our approach is more careful: we decompose entire texts rather than individual sentences to reduce redundancy and ambiguity, and we produce decontextualized claims to simplify verification. Claim quality was further examined during manual validation in complex cases.

\begin{figure}[h!]
    \begin{minipage}[t]{0.45\textwidth}
    \centering
    \fbox{
        \parbox{1.0\textwidth}{\scriptsize
            Contents (not necessarily includes answer to the following question):\\
            Title: \{title1\} \\
            Content: \{retrieval1\}\\
            ...\\
            Title: \{title5\} \\
            Content: \{retrieval5\}\\
            Question: \{question\}\\
            Answer (single line):
        }
    }
    \vspace{-0.5em}
    \caption{Prompt used in short-form QA datasets. Titles and retrievals correspond to the Wikipedia page title and the passage retrieved from it.}
    \label{fig:qa_prompt}
    \end{minipage}
    
      \vspace{1em}
    
    \begin{minipage}[t]{0.45\textwidth}
    \centering
    \fbox{
        \parbox{1.0\textwidth}{\scriptsize
            Using the context provided below, answer the question with a balanced approach. Ensure your response contains an equal number of claims or details drawn directly from the context and from your own knowledge:\\
            Context: passage 1:\{retrieval1\}\\
            passage 2:\{retrieval2\}\\
            passage 3:\{retrieval3\}\\
            Question: \{question\}\\
            Answer:
        }
    }
    \vspace{-0.5em}
    \caption{Prompt used in long-form QA datasets. Retrievals corresponds to the Wikipedia passage retrieved for input question.}
    \label{fig:claim_prompt}
\end{minipage}
\end{figure}

  \begin{table*}
\resizebox{1.0\textwidth}{!}{
\begin{tabular}{ll|cc|ccc|c}
\toprule
\textbf{Model} & \textbf{Dataset} & \textbf{Train Size} & \textbf{Test Size} & \textbf{True} & \textbf{False} & \textbf{Unverifiable} & \textbf{\multirowcell{Mean Generation\\ Length (characters)}} \\
\midrule
\multirow{4}{*}{\llama} & NQ & 200 & 1000 & 62.4 \% & 27.6 \% & 10.0 \% & 180.1 \\
 & PopQA & 200 & 1000 & 50.2 \% & 22.4 \% & 27.3 \% & 149.2 \\
 & TriviaQA & 200 & 1000 & 68.0 \% & 22.3 \% & 9.7 \% & 114.4 \\
 & SimpleQA & 200 & 1000 & 29.5 \% & 14.4 \% & 56.1 \% & 159.9 \\
 \midrule
\multirow{4}{*}{\falcon} & NQ & 200 & 1000 & 44.1 \% & 37.6 \% & 18.3 \% & 352.2 \\
 & PopQA & 200 & 1000 & 42.6 \% & 41.6 \% & 15.8 \% & 260.3 \\
 & TriviaQA & 200 & 1000 & 57.2 \% & 32.8 \% & 10.0 \% & 324.7 \\
 & SimpleQA & 200 & 1000 & 25.5 \% & 65.6 \% & 8.8 \% & 286.9 \\
\bottomrule
\end{tabular}
}
\caption{Statistics of datasets used in short-form QA benchmark.}
\label{tab:stats_sequence}
\end{table*}

  \begin{table*}
\resizebox{1.0\textwidth}{!}{
\begin{tabular}{l|cc|ccc|ccc|c}
\toprule
\textbf{Model} & \textbf{Train Size} & \textbf{Test Size} & \textbf{True} & \textbf{False} & \textbf{Unverifiable} & \textbf{Faithful} & \textbf{Unfaithful} & \textbf{Undefined} & \textbf{\multirowcell{Mean Generation\\ Length (characters)}} \\
\midrule
\llama & 600 & 1182 & 91.0 \% & 5.8 \% & 3.1 \% & 37.3 \% & 62.6 \% & 0.1 \% & 1725.4 \\
\falcon & 600 & 948 & 91.4 \% & 6.0 \% & 2.6 \% & 38.2 \% & 61.5 \% & 0.3 \% & 1720.2 \\
\llamalarge & 300 & 500 & 89.4 \% & 5.0 \% & 5.6 \% & 34.6 \% & 64.6 \% & 0.8 \% & 1856.4 \\
\gemmasmall & 300 & 500 & 88.8 \% & 5.7 \% & 5.5 \% & 44.7 \% & 54.5 \% & 0.8 \% & 1708.3 \\
\bottomrule
\end{tabular}
}
\caption{Statistics of datasets used in long-form QA benchmark.}
\label{tab:stats_claim}
\end{table*}

  \begin{figure}
      \begin{minipage}[t]{0.45\textwidth}
        \input{tables/claim_extraction_prompt}
      \end{minipage}
      \begin{minipage}[t]{0.45\textwidth}
        \input{tables/claim_matching_prompt}
      \end{minipage}
  \end{figure}

\newpage
\clearpage

  \begin{table}[t]
    \centering
    \small
    \begin{tabular}{p{0.17\linewidth} p{0.75\linewidth}}
      \toprule
      \textbf{Label} & \textbf{Instruction} \\
      \midrule
      \textbf{True} &
      Assign \textit{True} if the claim is supported by reliable, verifiable sources under its most natural interpretation. Minor wording differences are acceptable as long as the factual content is preserved. \\[0.4em]

      \textbf{False} &
      Assign \textit{False} if the claim is contradicted by reliable sources or contains an incorrect factual statement. \\[0.4em]

      \textbf{Unverifiable} &
      Assign \textit{Unverifiable} if the claim cannot be reliably confirmed or refuted from available sources, or if it is too ambiguous, underspecified, subjective, or dependent on missing context. \\
      \bottomrule
    \end{tabular}
    \caption{Protocol used for manual factuality verification of atomic claims.}
    \label{tab:annotation_guidelines}
  \end{table}

  \begin{table}
\centering
\resizebox{0.45\textwidth}{!}{
\begin{tabular}{lccc}
\toprule
\textbf{Annotation Type} & \textbf{Num of Claims} & \textbf{Accuracy} & \textbf{Cohen's Kappa} \\
\midrule
Factuality   & 100 & .87 & .552 \\
Faithfulness & 100 & .78 & .586 \\
\bottomrule
\end{tabular}
}
\caption{Inter-annotator agreement for factuality and faithfulness annotations based on 100 claims of \llama. Accuracy measures raw agreement, Cohen's Kappa adjusts for chance agreement.}
\label{tab:annotation_agreement}
\end{table}

\subsection{Long-Form Dataset Annotation}
\label{sec:longform_annotation}

    We annotate the long-form QA dataset using a two-stage pipeline. In the first stage, \gpto-search assigns faithfulness and factuality labels to extracted atomic claims using prompts. In the second stage, we manually verify the most difficult claims, namely those labeled as \textit{False} or \textit{Unverifiable}, and correct them when necessary. This design scales annotation while focusing human effort on cases where automatic labeling is least reliable.

\paragraph{Automatic First-Pass Annotation.}
\label{sec:annotation}

  Each claim is automatically annotated for both faithfulness and factuality. For faithfulness, \gpto-search assigns one of three labels: \textit{faithful}, \textit{unfaithful-contra}, or \textit{unfaithful-neutral}. In the experiments, these labels are binarized as faithful $\rightarrow 1$ and unfaithful-contra / unfaithful-neutral $\rightarrow 0$, since the unfaithful-contra class constitutes less than 5\% of the data.

  For factuality, \gpto-search assigns one of three labels: \textit{True}, \textit{False}, or \textit{Unverifiable}. Factuality is evaluated independently of the retrieved RAG context: the goal is to determine whether the claim is correct with respect to external knowledge rather than whether it is supported by the retrieved passages. For downstream evaluation, we retain only verifiable claims and binarize the labels as False $\rightarrow 1$ and True $\rightarrow 0$. The prompt used for automatic annotation is shown in Figure~\ref{fig:claim_anno_prompt}.

\paragraph{Targeted Manual Verification.}
\label{sec:manual_annotation}

\begin{figure*}
    \centering
    \begin{subfigure}{0.49\linewidth}
        \centering
        \includegraphics[width=\linewidth]{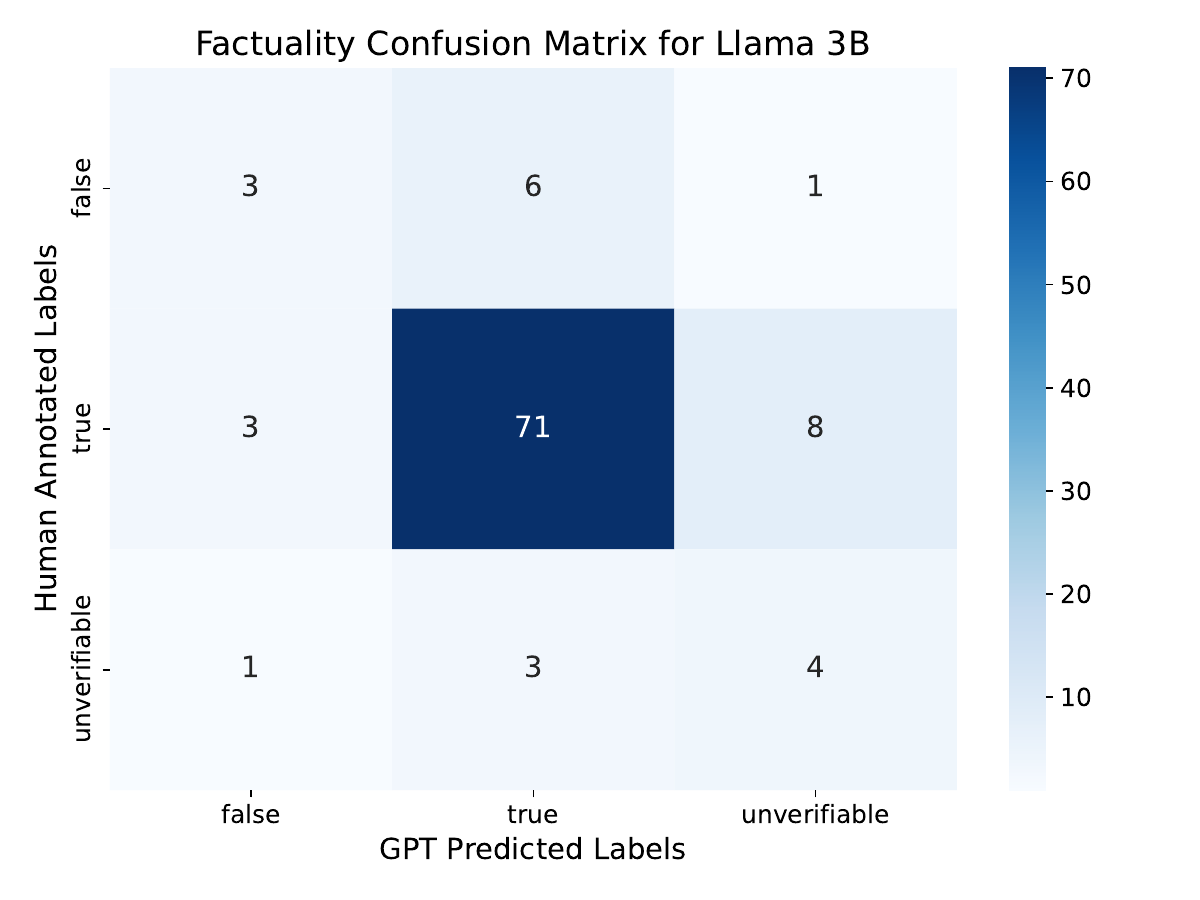}
        \caption{Before manual enhancement of automatic annotation}
    \end{subfigure}
    \hfill
    \begin{subfigure}{0.49\linewidth}
        \centering
        \includegraphics[width=\linewidth]{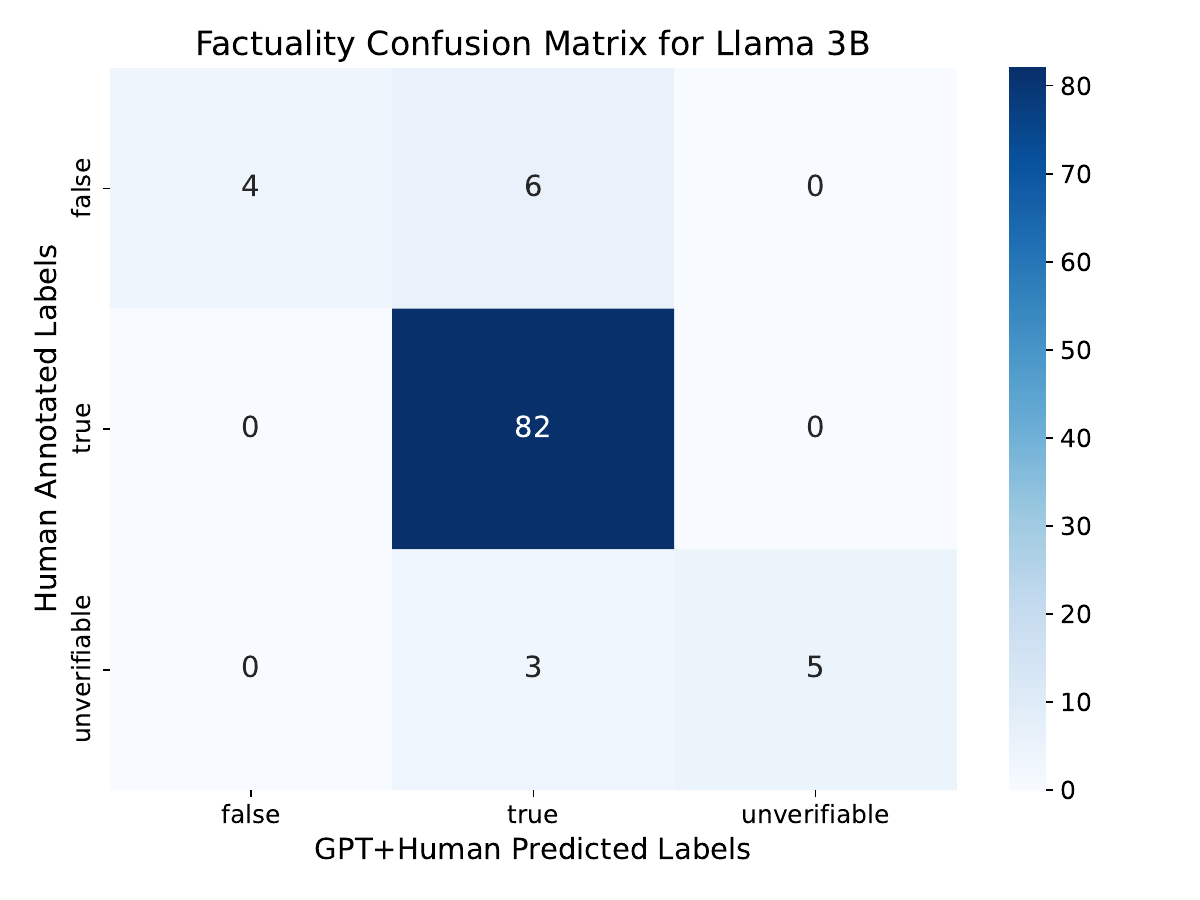}
        \caption{After manual enhancement of automatic annotation}
    \end{subfigure}
    \caption{Balance of classes of factuality annotations for the \llama model. Each matrix is based on 100 randomly selected claims, comparing annotations produced by the model with those from human annotators.}
  \label{fig:llama_factuality}
  \end{figure*}

\begin{figure*}
    \centering
    \begin{subfigure}{0.49\linewidth}
        \centering
        \includegraphics[width=\linewidth]{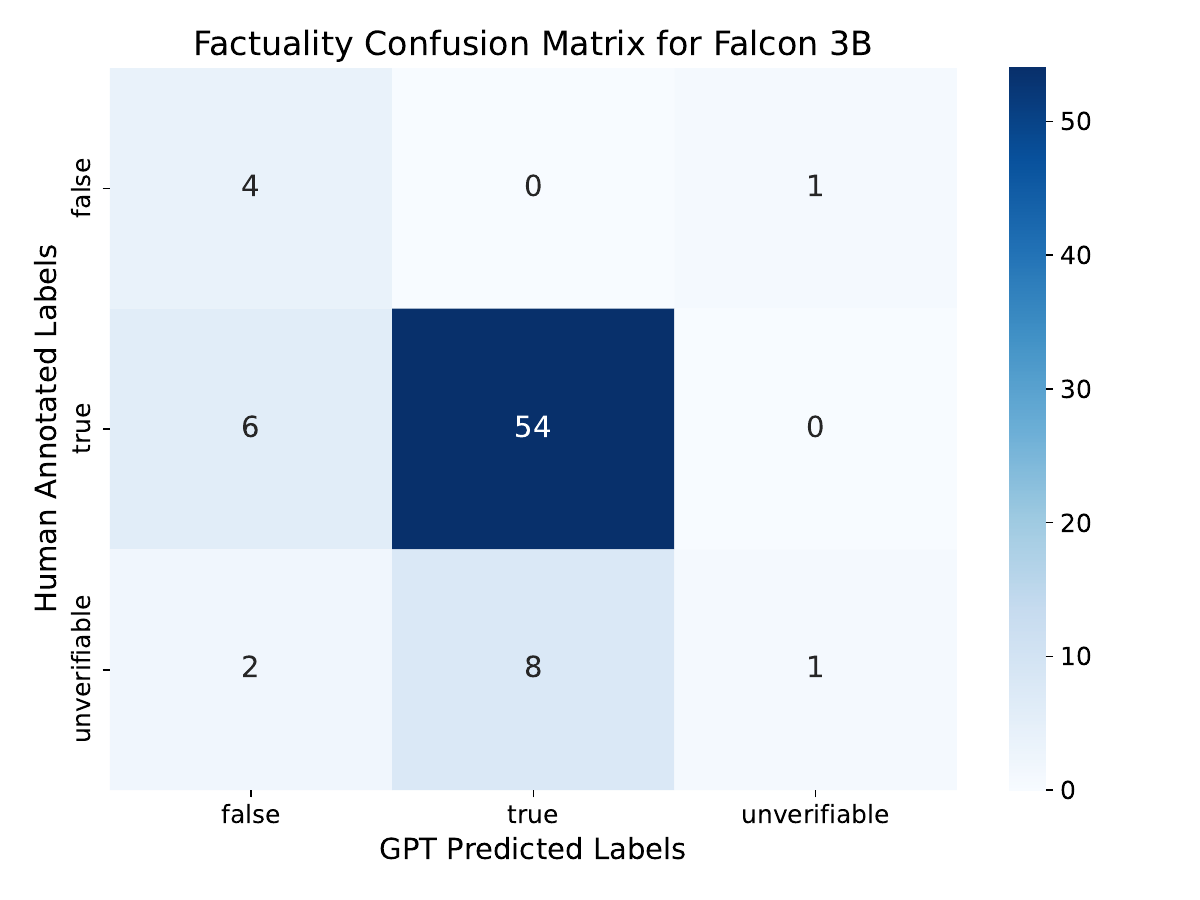}
        \caption{Before manual enhancement of automatic annotation}
    \end{subfigure}
    \hfill
    \begin{subfigure}{0.49\linewidth}
        \centering
        \includegraphics[width=\linewidth]{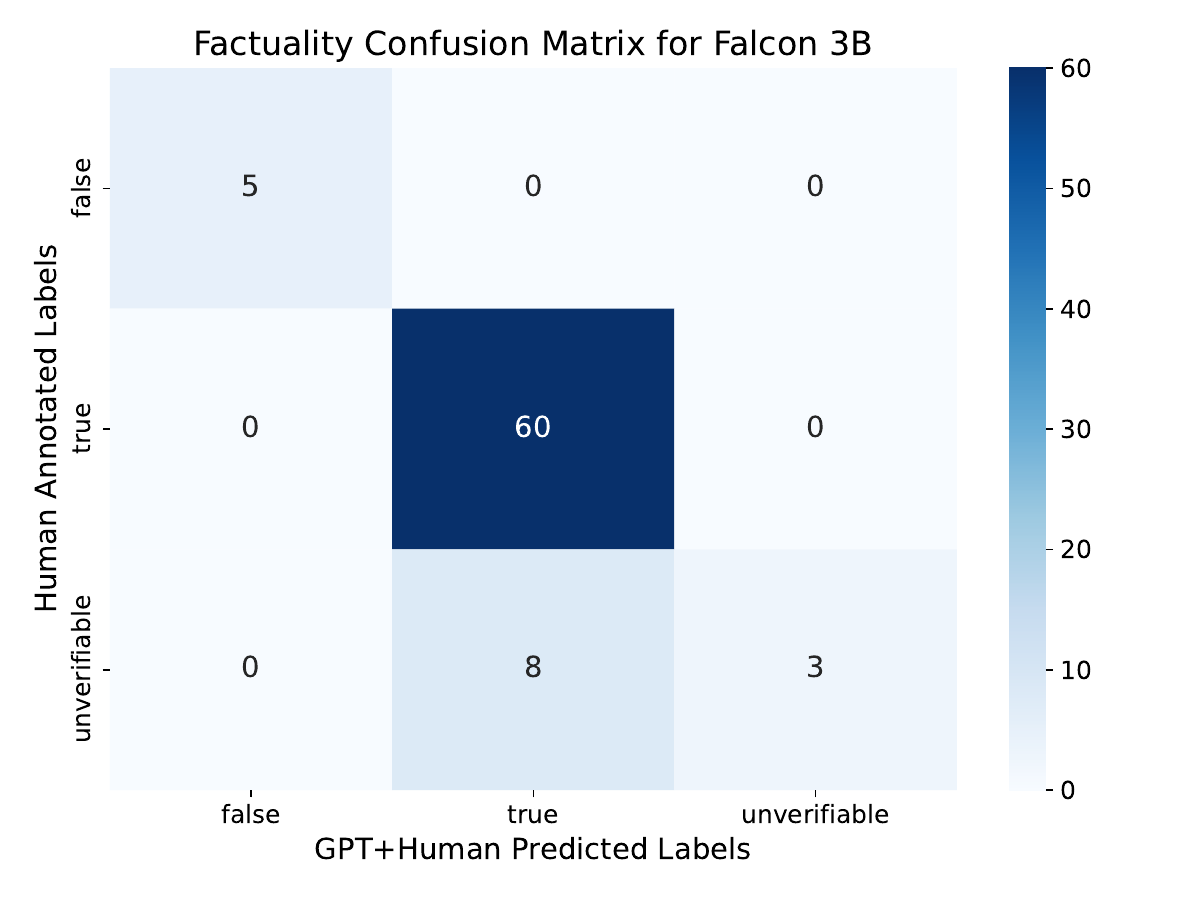}
        \caption{After manual enhancement of automatic annotation}
    \end{subfigure}
    \caption{Balance of classes of factuality annotations for the \falcon model. Each matrix is based on 76 randomly selected claims, comparing annotations produced by the model with those from human annotators.}
    \label{fig:falcon_factuality}
    \end{figure*}

\begin{figure*}
    \centering
    \begin{subfigure}{0.49\linewidth}
        \centering
        \includegraphics[width=\linewidth]{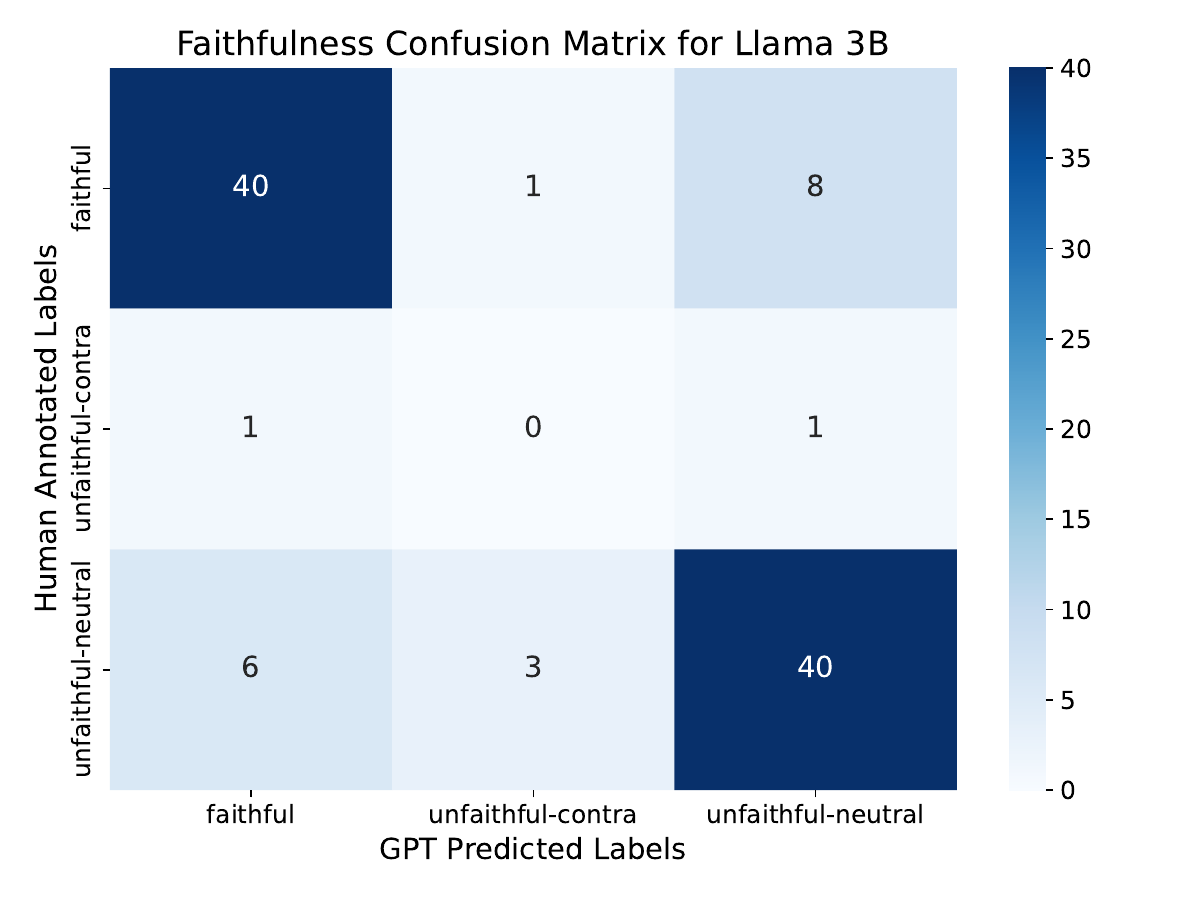}
        \caption{\llama faithfulness classes}
    \end{subfigure}
    \hfill
    \begin{subfigure}{0.49\linewidth}
        \centering
        \includegraphics[width=\linewidth]{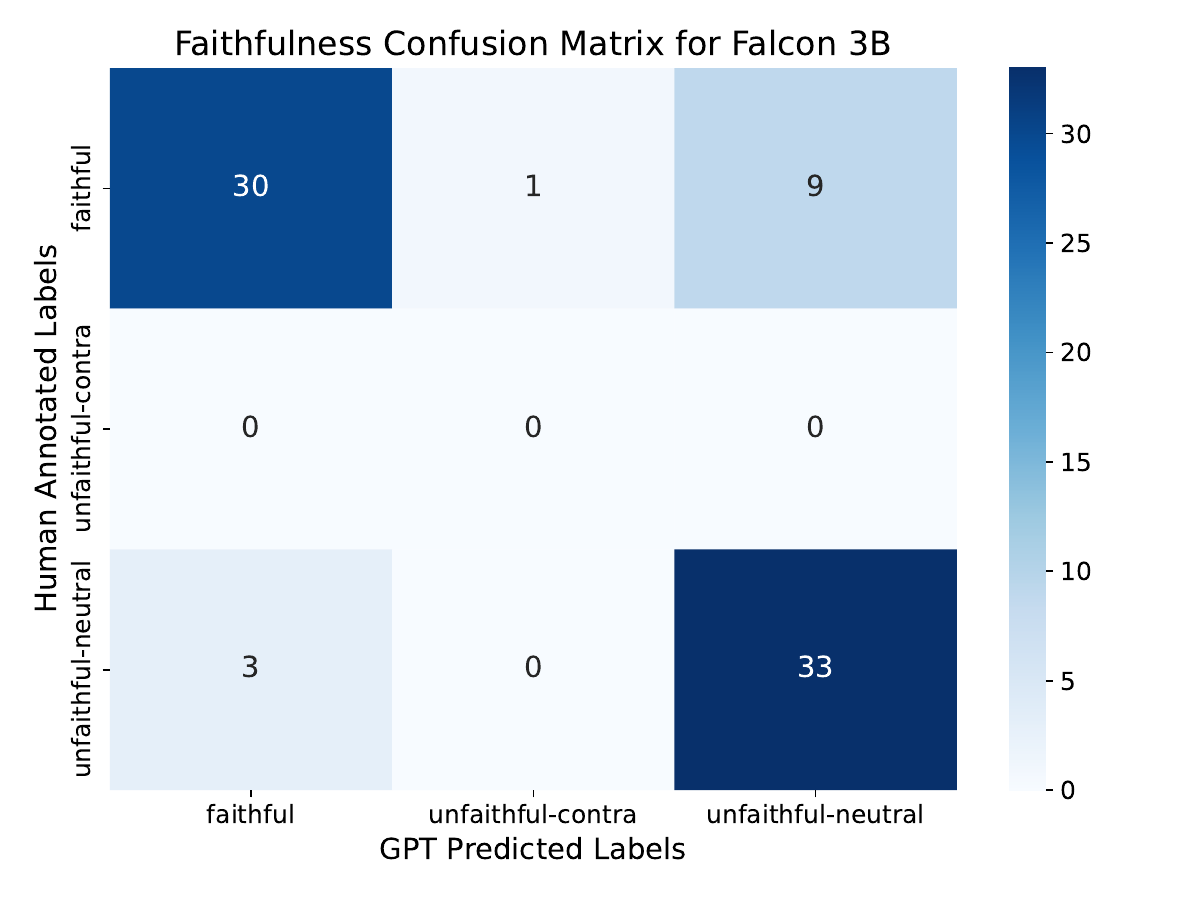}
        \caption{\falcon faithfulness classes}
    \end{subfigure}
    \caption{Balance of classes of faithfulness annotations for \llama and \falcon models. The matrices are based on 100 and 76 randomly selected claims, correspondingly, comparing annotations produced by the model with those from human annotators.}
  \label{fig:faithfulness}
  \end{figure*}

    We manually validate the automatic annotations to assess their quality and to correct the most difficult cases. First, to estimate automatic annotation reliability and class balance, we compare automatic and human labels on randomly selected claims: 100 for \llama and 76 for \falcon. The resulting factuality comparisons are shown in Figure~\ref{fig:llama_factuality}(a) and Figure~\ref{fig:falcon_factuality}(a); corresponding faithfulness comparisons are shown in Figure~\ref{fig:faithfulness}(a, b), providing a quantitative assessment of annotation accuracy and consistency across both evaluated models.

  Our validation shows that factuality labels predicted as \textit{False} or \textit{Unverifiable} are substantially less reliable than those predicted as \textit{True}. We therefore use a targeted second stage in which all claims initially labeled as \textit{False} or \textit{Unverifiable} are manually re-checked using reliable external sources identified through web search. This yields 359 manually reviewed claims for \llama and 240 for \falcon. The resulting corrected label distributions for the sampled subsets are shown in Figure~\ref{fig:llama_factuality}(b) and Figure~\ref{fig:falcon_factuality}(b).

  Six student annotators contributed to this manual verification stage, each spending about three hours on the task. Annotators followed the simple factuality protocol summarized in Table~\ref{tab:annotation_guidelines}. All annotators volunteered and received no financial compensation.

    To evaluate annotation consistency, we additionally conduct an agreement analysis on the 100 \llama claims, each independently reviewed by two annotators (Table~\ref{tab:annotation_agreement}). The results indicate generally strong agreement for factuality, suggesting that the annotation process is reliable and consistent across annotators, and that disagreements are relatively rare and limited in scope.
    
  \begin{figure}
    \centering
    \fbox{
        \parbox{0.45\textwidth}{\scriptsize
Evaluate the given claim using two criteria: \textbf{faithfulness} and \textbf{factuality}.\\
- \textbf{Faithfulness} assesses how accurately the claim reflects the \textit{context document}. Assign one of the following labels:\\
  - "faithful" — The claim is directly supported by the context.\\
  - "unfaithful-contra" — The claim directly contradicts the context.\\
  - "unfaithful-neutral" — The claim is not supported by the context.\\
- \textbf{Factuality} assesses the truth of the claim \textit{independently of the context}, based on the most up-to-date and reliable sources of knowledge available to humanity. Assign one of the following labels:\\
  - "True" — The claim is factually correct.\\
  - "False" — The claim is factually incorrect.\\
  - "unverifiable" — The truth cannot be determined with current knowledge.\\
Return your answer in the exact format:
("faith. label", "factuality label")\\
Context Document: \{retrievals\}  \\
Claim: \{claim\}  \\
Label:
        }
    }
    \caption{Prompt used with \gpto-search to automatically annotate claims for faithfulness and factuality in long-form QA benchmark. }
    \label{fig:claim_anno_prompt}
\end{figure}


\newpage
\clearpage

    \begin{figure}[tbh]
    \centering
    \fbox{
    \begin{minipage}{0.95\linewidth}
    \small
    
    \textbf{Question:} \textit{How and when to harvest chestnuts?}
    
    \vspace{0.3em}
    \textbf{Retrieved passage (excerpt):}  
    ``When to harvest chestnuts? Chestnuts don’t all ripen at once. Harvest typically spans up to five weeks, but most nuts ripen within a 10--30 day period in late August and September.''
    
    \vspace{0.3em}
    \textbf{Model-generated claim:}  
    ``The best time to harvest chestnuts is during the 10--30 day ripening window.''
    
    \vspace{0.3em}
    \textbf{AlignScore:} 0.61
    
    \vspace{0.3em}
    \textbf{Analysis:}  
    While the retrieved passage mentions a 10--30 day ripening period, it does not specify that this window constitutes the \emph{best} time for harvesting. Accordingly, AlignScore assigns this claim an intermediate faithfulness score of 0.61, reflecting partial grounding.
    \end{minipage}
    }
    \caption{Example illustrating intermediate AlignScore values arising from partial grounding between a claim and retrieved evidence.}
    \label{fig:faithfulness_example}
    \end{figure}

\section{Additional Faithfulness-Related Analysis}
\label{section:appendix_faithfulness}

    In this section, we provide additional analysis of the behavior and effectiveness of AlignScore as a faithfulness estimator within the FRANQ framework, examining its performance and robustness across different datasets and settings.

\subsection{Faithfulness Distribution and Calibration}

\begin{figure*}
    \centering
    \includegraphics[width=\linewidth]{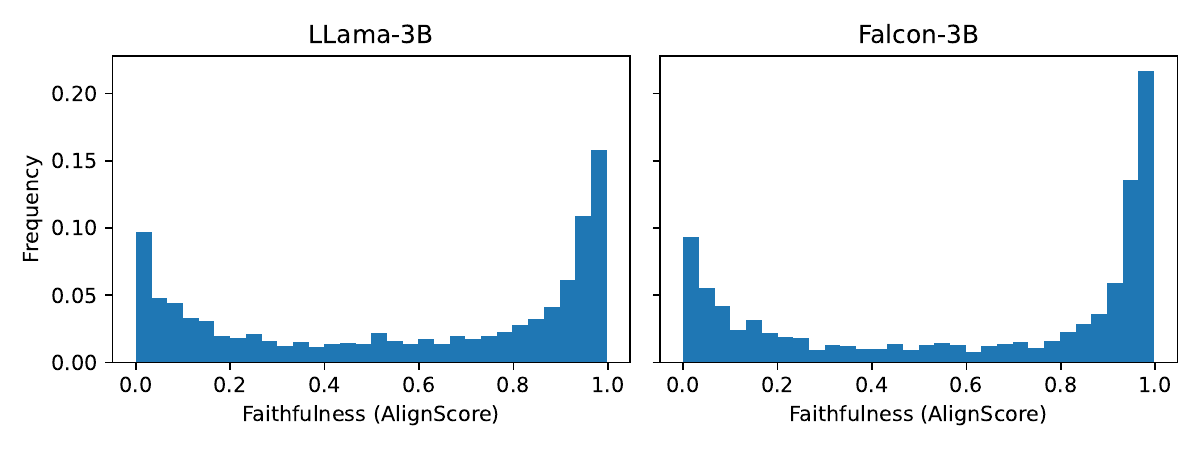}
    \caption{Distribution of AlignScore-based faithfulness estimates on the long-form QA benchmark for \llama and \falcon. A substantial mass lies in the intermediate region, and low ECE values indicate good calibration.}
    \label{fig:faithfulness_distribution}
  \end{figure*}

    We examine the empirical behavior of AlignScore when used to estimate claim-level faithfulness. Figure~\ref{fig:faithfulness_distribution} shows the distribution of AlignScore values computed between model-generated claims and their corresponding retrieved documents on the long-form QA benchmark, for two representative models, providing insight into how faithfulness scores are distributed across different claims.

  We observe that a substantial fraction of claims receive intermediate faithfulness scores, reflecting cases where claims are only partially supported or rely on implicit inferences from the retrieved evidence. Across both models, more than 40\% of claims fall within the range $[0.1, 0.9]$.

    AlignScore also demonstrates strong calibration with respect to gold faithfulness labels, achieving low expected calibration error (ECE $= 0.05$). This indicates that AlignScore provides a reliable continuous estimate of faithfulness suitable for probabilistic combination in \franq equation~\ref{eq:cuq}, and can be effectively integrated with other uncertainty estimation components.

    Figure~\ref{fig:faithfulness_example} provides a representative qualitative example illustrating how intermediate faithfulness values arise in practice.

\subsection{Faithfulness Evaluation on Long-Form QA}
  \begin{table*}[ht]
    \centering
    \begin{subtable}[t]{0.48\textwidth}
      \centering
\resizebox{\textwidth}{!}{
\begin{tabular}{l|ccc}
 \toprule
  \multirow{2}{*}{\textbf{Method}} & \multicolumn{3}{c}{\textbf{\llama}} \\
 & \textbf{AUROC $\uparrow$} & \textbf{PR-AUC $\uparrow$} & \textbf{PRR $\uparrow$} \\
\midrule
 \rowcolor{gray!20}
 \multicolumn{4}{c}{General Baselines} \\
\midrule
Max Claim Prob. & .614 & .751 & .298 \\
P(True) & .447 & .624 & -.242 \\
Perplexity & \underline{.642} & \underline{.782} & \underline{.315} \\
Mean Token Entropy & .596 & .743 & .208 \\
CCP & .569 & .727 & .135 \\
\midrule
 \rowcolor{gray!20}
 \multicolumn{4}{c}{RAG-Specific Baselines} \\
\midrule
AlignScore & \textbf{.856} & \textbf{.907} & \textbf{.789} \\
Parametric Knowledge & .273 & .559 & -.704 \\
\bottomrule
\end{tabular}
}
\caption{Results on long-form QA benchmark with faithfulness target.}
\label{tab:claim_faithfulness}

    \end{subtable}
    \hfill
    \begin{subtable}[t]{0.48\textwidth}
      \centering
\resizebox{\textwidth}{!}{
\begin{tabular}{l|ccc}
 \toprule
  \multirow{2}{*}{\textbf{Method}} & \multicolumn{3}{c}{\textbf{\llama}} \\
 & \textbf{AUROC $\uparrow$} & \textbf{PR-AUC $\uparrow$} & \textbf{PRR $\uparrow$} \\
\midrule
 \rowcolor{gray!20}
 \multicolumn{4}{c}{General Baselines} \\
\midrule
Max Claim Prob. & .538 & .115 & .028 \\
P(True) & .463 & .112 & .002 \\
Perplexity & .480 & .092 & -.068 \\
Mean Token Entropy & .580 & \underline{.167} & .122 \\
CCP & \underline{.585} & .134 & \underline{.152} \\
\midrule
 \rowcolor{gray!20}
 \multicolumn{4}{c}{RAG-specific Baselines} \\
\midrule
AlignScore & .477 & .094 & -.007 \\
Parametric Knowledge & \textbf{.667} & \textbf{.190} & \textbf{.303} \\
\bottomrule
\end{tabular}
}
\caption{Results on long-form QA benchmark with factuality target (only unfaithful claims).}
\label{tab:claim_factuality_if_unfaithful}

    \end{subtable}
    \caption{Additional faithfulness-related results}
  \end{table*}

We next evaluate the effectiveness of AlignScore as a faithfulness estimator on the long-form QA benchmark. Table~\ref{tab:claim_faithfulness} reports performance when faithfulness is treated as the target metric. All methods follow the same experimental setup used for the factuality evaluation, ensuring a fair and consistent comparison across approaches and settings, and isolating the contribution of each estimator to faithfulness prediction performance under controlled conditions and identical input configurations.

Among the compared approaches, AlignScore achieves the strongest performance across metrics, indicating its effectiveness in approximating claim-level faithfulness within the FRANQ decomposition and supporting its suitability as a core component of the framework for reliable faithfulness estimation, particularly in long-form generation settings with multiple claims, diverse evidence sources, and complex reasoning requirements.

\subsection{Factuality Under Faithful and Unfaithful Conditions}
\label{section:appendix_fact_if_faith_or_unfaith}

We further analyze factuality estimation under faithful and unfaithful conditions. Table~\ref{tab:claim_factuality_if_unfaithful} reports results for unsupervised methods when restricting evaluation to unfaithful claims only. In this setting, methods leveraging parametric knowledge perform best, achieving the highest AUROC and PRR scores, highlighting the importance of modeling internal knowledge when retrieval grounding is unreliable and external evidence may be misleading or partially incorrect in realistic scenarios.

Table~\ref{tab:qa_meanrank_faithful_unfaithful} report results averaged across four QA datasets for \llama, considering only claims with high and low AlignScore, respectively, to separate faithful and unfaithful regimes and analyze performance differences more precisely across datasets and evaluation conditions. For faithful claims, Semantic Entropy achieves the best performance, whereas for unfaithful claims, the sum of eigenvalues of the Graph Laplacian performs best. These results further motivate the use of different uncertainty estimators conditioned on faithfulness within FRANQ, demonstrating the benefit of tailoring uncertainty estimation to distinct error sources and improving robustness across diverse retrieval conditions, dataset characteristics, varying levels of evidence quality, and different model behaviors, as well as across varying levels of task difficulty.

  \begin{table*}[ht]
    \centering
    \begin{subtable}[t]{0.48\textwidth}
      \centering

\resizebox{\textwidth}{!}{
\begin{tabular}{l|rrr}
\toprule
 \multirow{2}{*}{\textbf{Method}} 
 & \multicolumn{3}{c}{\textbf{\llama}} \\
 & \textbf{AUROC $\uparrow$} & \textbf{PR-AUC $\uparrow$} & \textbf{PRR $\uparrow$} \\
\midrule
 \rowcolor{gray!20}
 \multicolumn{4}{c}{General Baselines} \\
\midrule
Max Sequence Prob. & .754 & .518 & .454 \\
Mean Token Entropy & .767 & .540 & .472 \\
CCP & .742 & .512 & .434 \\
Lexical Similarity & .758 & .500 & .457 \\
Degree Matrix & \underline{.770} & \underline{.549} & \underline{.488} \\
Sum of Eigenvalues & .767 & .538 & .476 \\
Semantic Entropy & \textbf{.781} & \textbf{.562} & \textbf{.510} \\
SentenceSAR & .766 & .518 & .473 \\
\midrule
 \rowcolor{gray!20}
 \multicolumn{4}{c}{RAG-Specific Baselines} \\
\midrule
AlignScore & .606 & .321 & .170 \\
Parametric Knowledge & .657 & .413 & .295 \\
\bottomrule
\end{tabular}
}
\caption{Only claims with AlignScore > 0.5}
\label{tab:qa_meanrank_llama_if_faithful}

    \end{subtable}
    \hfill
    \begin{subtable}[t]{0.48\textwidth}
      \centering

\resizebox{\textwidth}{!}{
\begin{tabular}{l|rrr}
\toprule
 \multirow{2}{*}{\textbf{Method}} 
 & \multicolumn{3}{c}{\textbf{\llama}} \\
 & \textbf{AUROC $\uparrow$} & \textbf{PR-AUC $\uparrow$} & \textbf{PRR $\uparrow$} \\
\midrule
 \rowcolor{gray!20}
 \multicolumn{4}{c}{General Baselines} \\
\midrule
Max Sequence Prob. & .752 & .648 & .446 \\
Mean Token Entropy & .755 & .673 & .462 \\
CCP & .741 & .631 & .445 \\
Lexical Similarity & .767 & .662 & .469 \\
Degree Matrix & \underline{.796} & \underline{.728} & \underline{.551} \\
Sum of Eigenvalues & \textbf{.807} & \textbf{.735} & \textbf{.560} \\
Semantic Entropy & .782 & .689 & .502 \\
SentenceSAR & .770 & .667 & .473 \\
\midrule
 \rowcolor{gray!20}
 \multicolumn{4}{c}{RAG-Specific Baselines} \\
\midrule
AlignScore & .555 & .488 & .142 \\
Parametric Knowledge & .602 & .512 & .230 \\
\bottomrule
\end{tabular}
}
\caption{Only claims with AlignScore < 0.5}
\label{tab:qa_meanrank_llama_if_unfaithful}

    \end{subtable}
    \caption{Results averaged across 4 QA datasets for \llama considering only claims with high and low AlignScore.}
    \label{tab:qa_meanrank_faithful_unfaithful}
  \end{table*}

\newpage
\clearpage
\newpage
\clearpage

\section{Additional Ablation Studies}
\label{sec:appendix_ablations}

  \begin{table}[h!]
\centering
\resizebox{1.0\linewidth}{!}{

\begin{tabular}{l|lll}
\toprule
\multirow{2}{*}{\textbf{Method}} & \multicolumn{3}{c}{\textbf{\llama, long-form QA}} \\
 & \textbf{AUROC $\uparrow$} & \textbf{PR-AUC $\uparrow$} & \textbf{PRR $\uparrow$} \\
\midrule
\franq no calibration & \underline{.646} & .100 & .181 \\
\franq calibrated & \textbf{.653} & .103 & \textbf{.256} \\
\franq condition-calibrated & .641 & \textbf{.140} & \underline{.223} \\
\franq condition-, faithfulness-calibrated & .587 & \underline{.124} & .112 \\
\bottomrule
\end{tabular}

}
\caption{Comparison of \franq performance on \llama long-form QA benchmark, when applying calibration for faithfulness estimator, AlignScore.}
\label{tab:ablations_alignscore_calibration}
\end{table}
  
\subsection{\franq with Alternative Faithfulness Estimators}
\label{sec:ablations_thresholded_faith}

Table~\ref{tab:ablations_thresholded} compares the performance of three original \franq versions (each employing a different calibration strategy) with three modified versions that use a thresholded AlignScore instead of raw AlignScore probabilities. In the thresholded versions, the faithfulness probability is defined as $P(c\text{ is faithful to \retr}) = \mathds{1}\left(\text{AlignScore}(c) > T\right)$ with $T=0.5$. These methods are denoted by the `T=0.5' label. The results indicate that, overall, the continuous versions of \franq outperform their thresholded counterparts, suggesting that preserving the full probabilistic signal of AlignScore is important for effective uncertainty estimation and leads to better integration within the FRANQ decomposition. In particular, thresholding discards intermediate confidence values that may capture partial grounding or uncertainty, which can be informative for downstream factuality estimation and improve robustness in borderline cases, especially when evidence is incomplete or partially relevant.

Table~\ref{tab:ablations_alignscore_calibration} further compares the performance of three original \franq versions with a condition-calibrated version of \franq that also calibrates AlignScore for faithfulness estimation (this method is denoted `\franq condition-calibrated, faithfulness-calibrated'). In this version, the AlignScore is calibrated using a training set with binary gold faithfulness targets and then incorporated into the \franq formula. The results suggest that calibrating AlignScore may reduce the PRR of \franq, indicating that it might be more effective to use AlignScore without faithfulness calibration, possibly due to the loss of useful ranking information during calibration and reduced sensitivity to fine-grained differences between claims.

  \begin{table*}[!ht]
\centering
\resizebox{\textwidth}{!}{

\begin{tabular}{l|lll|lll}
\toprule
\multirow{2}{*}{\textbf{Method}} & \multicolumn{3}{c|}{\multirowcell{\textbf{\llama,} \\ \textbf{long-form QA}}} & \multicolumn{3}{c}{\multirowcell{\textbf{\llama} \\ \textbf{mean across 4 short-form QA}}} \\
 & \textbf{AUROC $\uparrow$} & \textbf{PR-AUC $\uparrow$} & \textbf{PRR $\uparrow$} & \textbf{AUROC $\uparrow$} & \textbf{PR-AUC $\uparrow$} & \textbf{PRR $\uparrow$} \\
\midrule
\franq no calibration & \underline{.646} & .100 & .181 & .728 & .553 & .403 \\
\franq no calibration T=0.5 & .629 & .105 & .170 & .782 & .599 & .524 \\
\midrule
\franq calibrated & \textbf{.653} & .103 & \textbf{.256} & \underline{.797} & \underline{.628} & \underline{.537} \\
\franq calibrated T=0.5 & .607 & .085 & .084 & .796 & .566 & .521 \\
\midrule
\franq condition-calibrated & .641 & \textbf{.140} & \underline{.223} & \textbf{.802} & \textbf{.631} & \textbf{.541} \\
\franq condition-calibrated T=0.5 & .587 & \underline{.111} & .180 & .793 & .556 & .515 \\
\bottomrule
\end{tabular}

}
\caption{Comparison of \franq performance on \llama benchmarks, when using AlignScore with and without threshold.}
\label{tab:ablations_thresholded}
\end{table*}

\subsection{Analysis of XGBoost}
We examine the first tree from an XGBoost model trained on \franq features (AlignScore, Claim Probability, and Parametric Knowledge) for long-form QA with \llama. While XGBoost uses multiple trees, the first tree often captures key decision patterns and provides an interpretable approximation of the model’s behavior.

Figure~\ref{fig:ablations_xgboost_tree} presents the first several nodes in the first XGBoost tree. The root splits on AlignScore. If it's high, the model next considers Claim Probability; if low, it turns to Parametric Knowledge. This mirrors \franq's logic: leading with faithfulness assessment with AlignScore, followed by either Claim Probability or Parametric Knowledge. The tree thus exhibits structure similar to \franq's decision process, highlighting alignment between learned and designed decision strategies.

  \begin{figure}
\centering
\resizebox{1.0\linewidth}{!}{%
\begin{forest}
for tree={
    draw,
    rounded corners,
    align=center,
    edge={->,>=latex},
    parent anchor=south,
    child anchor=north,
    s sep=8mm,
    l sep=12mm,
    font=\footnotesize
}
[AlignScore > 0.72
    [ClaimProb > 0.39, edge label={node[midway,xshift=-6pt]{true}}
        [\dots, edge label={node[midway,left]{true}}]
        [\dots, edge label={node[midway,right]{false}}]
    ]
    [Parametric Knowledge > exp(-22.13), edge label={node[midway,xshift=6pt]{false}}
        [\dots = -0.30, edge label={node[midway,left]{true}}]
        [\dots, edge label={node[midway,right]{false}}]
    ]
]
\end{forest}
}
\caption{Top vertices of first XGBoost tree trained on \franq components (ClaimProb) for long-form QA \llama behchmark.}
\label{fig:ablations_xgboost_tree}
\end{figure}


\subsection{Calibration Properties of UQ Methods}
  We evaluate the calibration properties of all our UQ methods using the Expected Calibration Error (ECE; \citealp{pmlr-v70-guo17a}). ECE quantifies the alignment between predicted confidence scores and observed accuracy. Specifically, predictions are partitioned into 10 equally spaced confidence bins. Within each bin, we compute the average predicted confidence and compare it to the empirical accuracy. Lower ECE values indicate better-calibrated models.

  Table~\ref{tab:ablations_ece} reports ECE scores for both long-form QA dataset and short-form QA benchmark using the \llama model. Only UQ methods that produce confidence values within the [0, 1] interval are included, as this is a prerequisite for ECE computation. Notably, the two calibrated variants of \franq achieve the best calibration performance across datasets.

  \begin{table*}[ht]
    \centering
    \begin{subtable}[t]{0.48\textwidth}
          \centering
\begin{tabular}{l|c}
\toprule
\textbf{Method} & \textbf{ECE $\downarrow$} \\
\midrule
 \rowcolor{gray!20}
 \multicolumn{2}{c}{General Baselines} \\
\midrule
Max Claim Prob. & .72 \\
P(True) & .94 \\
Perplexity & .18 \\
CCP & .21 \\
\midrule
 \rowcolor{gray!20}
 \multicolumn{2}{c}{RAG-Specific Baselines} \\
\midrule
AlignScore & .40 \\
Parametric Knowledge & .80 \\
\midrule
 \rowcolor{gray!20}
 \multicolumn{2}{c}{XGBoost} \\
\midrule
XGBoost (all UQ features) & .05 \\
XGBoost (\franq features) & .06 \\
\midrule
 \rowcolor{gray!20}
 \multicolumn{2}{c}{\franq} \\
\midrule
\franq no calibration & .44 \\
\franq calibrated & \textbf{.02} \\
\franq condition-calibrated & \underline{.03} \\
\bottomrule
\end{tabular}

          \caption{Long-form QA \llama dataset.}
    \end{subtable}
    \hfill
    \begin{subtable}[t]{0.48\textwidth}
          \centering
\begin{tabular}{l|c}
\toprule
\textbf{Method} & \textbf{Mean ECE $\downarrow$} \\
\midrule
 \rowcolor{gray!20}
 \multicolumn{2}{c}{General Baselines} \\
\midrule
Max Sequence Prob. & .46 \\
Lexical Similarity & \textbf{.07} \\
Degree Matrix & .14 \\
Sum of Eigenvalues & .54 \\
CCP & .23 \\
\midrule
 \rowcolor{gray!20}
 \multicolumn{2}{c}{RAG-Specific Baselines} \\
\midrule
AlignScore & \underline{.13} \\
Parametric Knowledge & .23 \\
\midrule
 \rowcolor{gray!20}
 \multicolumn{2}{c}{XGBoost} \\
\midrule
XGBoost (all UQ features) & .15 \\
XGBoost (\franq features) & .17 \\
\midrule
 \rowcolor{gray!20}
 \multicolumn{2}{c}{\franq} \\
\midrule
\franq no calibration & .64 \\
\franq calibrated & \textbf{.07} \\
\franq condition-calibrated & \textbf{.07} \\
\bottomrule
\end{tabular}


          \caption{Short-form QA \llama benchmark (ECE is averaged across 4 QA datasets).}
    \end{subtable}
    \caption{Expected Calibration Error (ECE) for all tested UQ methods with \llama.}
    \label{tab:ablations_ece}
  \end{table*}





\newpage
\clearpage
\newpage
\clearpage

\section{Resource and Expenses}
  A full data-generation and UQ-baseline evaluation run required about 8 days of compute on an NVIDIA V100 32GB GPU for long-form QA, while short-form QA needed under one day. The OpenAI API was used for claim splitting, matching, and annotation, costing roughly \$100 per model run (\llama). Human annotation involved six student annotators, each contributing about three hours of work.

\section{\franq Examples}
\label{sec:franq_examples}
\begin{figure*}[htbp]
    \centering
    \begin{subfigure}{\linewidth}
        \centering
        \includegraphics[width=.8\linewidth]{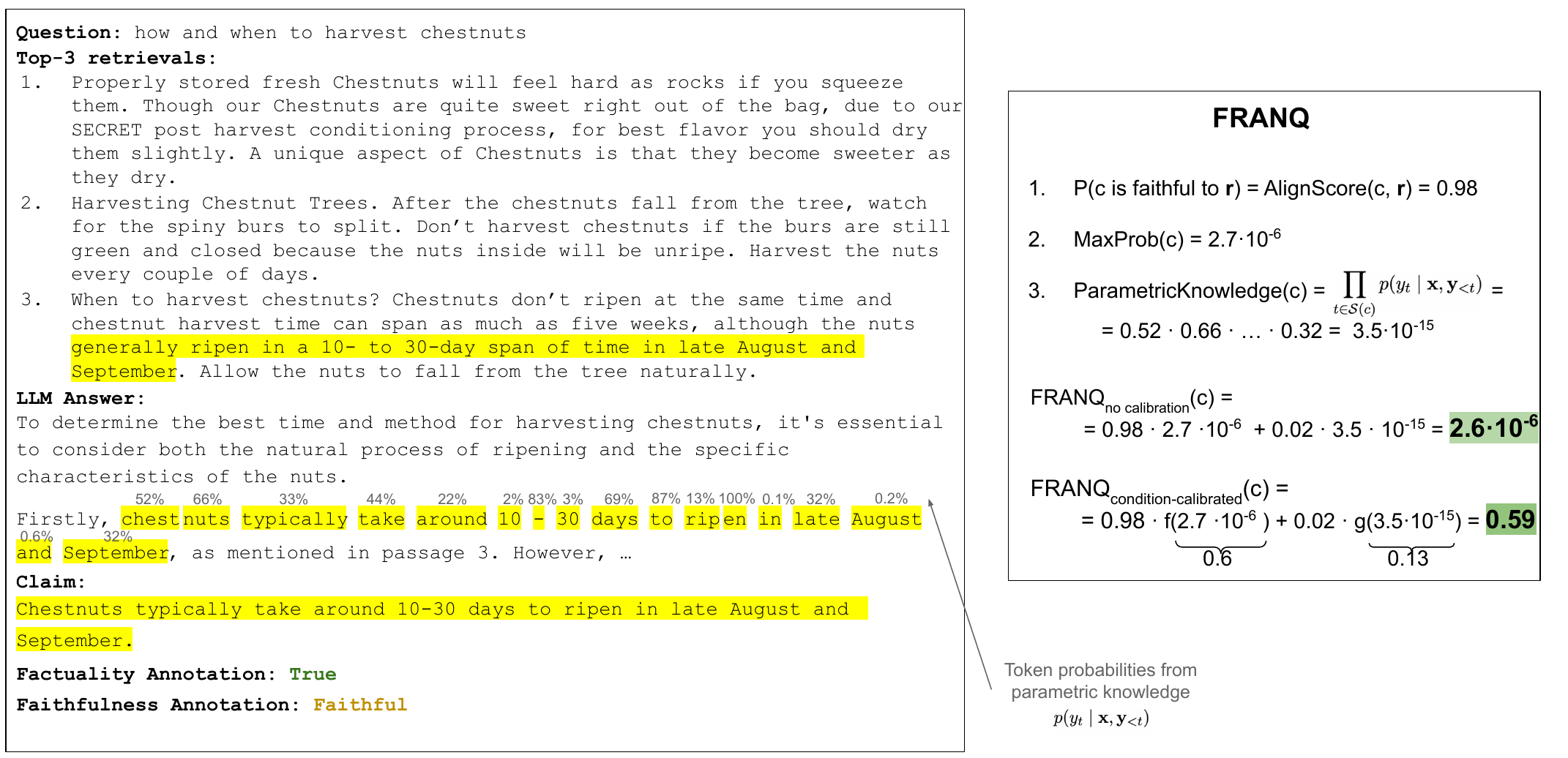}
        \caption{\textbf{Faithful–True.} \franq correctly identifies the claim as faithful and uses Claim Probability, which detects high entailment with the third retrieved passage. This results in an appropriately high \franq score.}
    \end{subfigure}
    \\
    \begin{subfigure}{\linewidth}
        \centering
        \includegraphics[width=.8\linewidth]{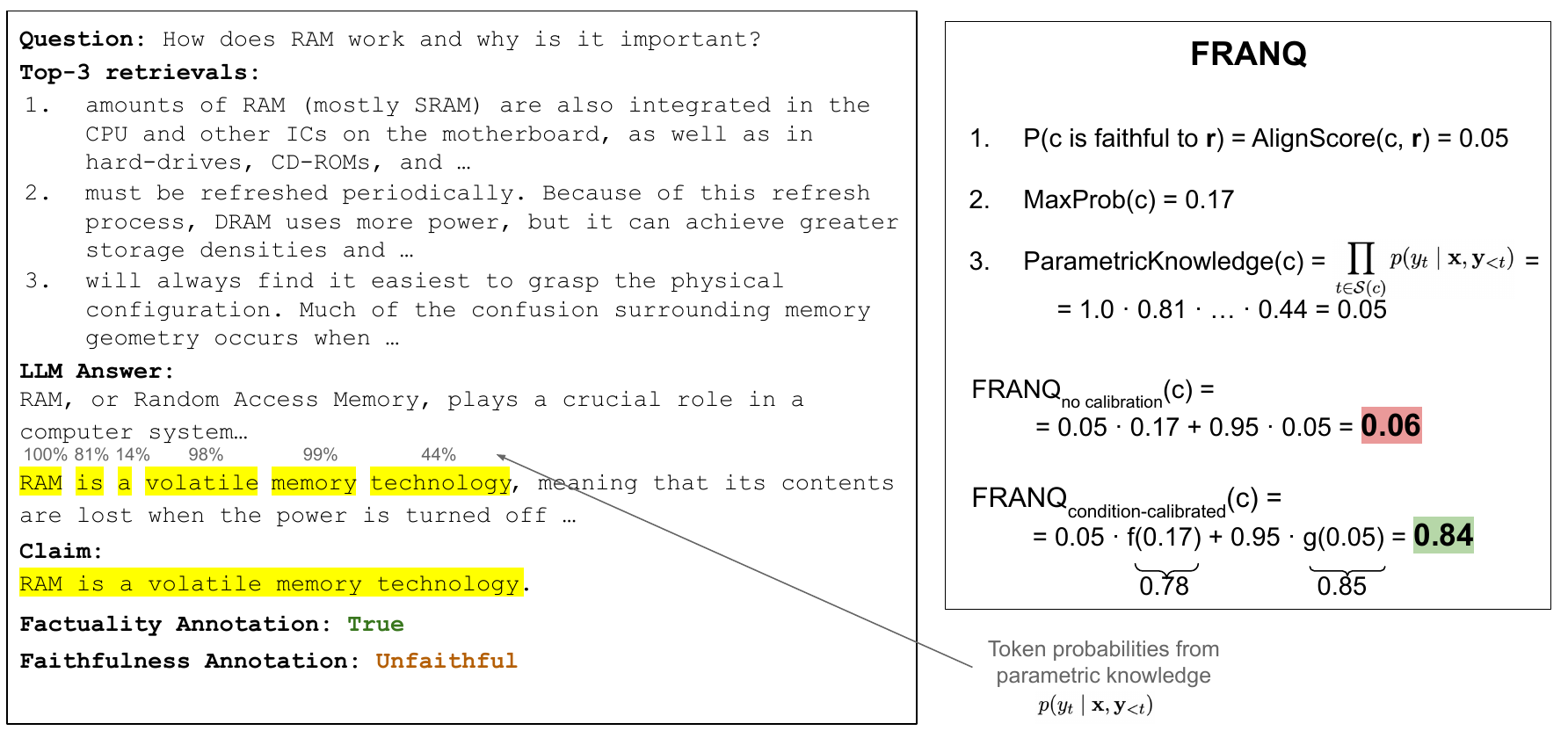}
        \caption{\textbf{Unfaithful–True.} \franq accurately detects the claim's low faithfulness and assigns its factuality score based on Parametric Knowledge, which is relatively high. In the uncalibrated version, the final score is underestimated due to the uncalibrated Parametric Knowledge score. The condition-calibrated version corrects this by assigning a calibrated score of 0.85, resulting in a correctly high factuality estimate.}
    \end{subfigure}
    \\
    \begin{subfigure}{\linewidth}
        \centering
        \includegraphics[width=.8\linewidth]{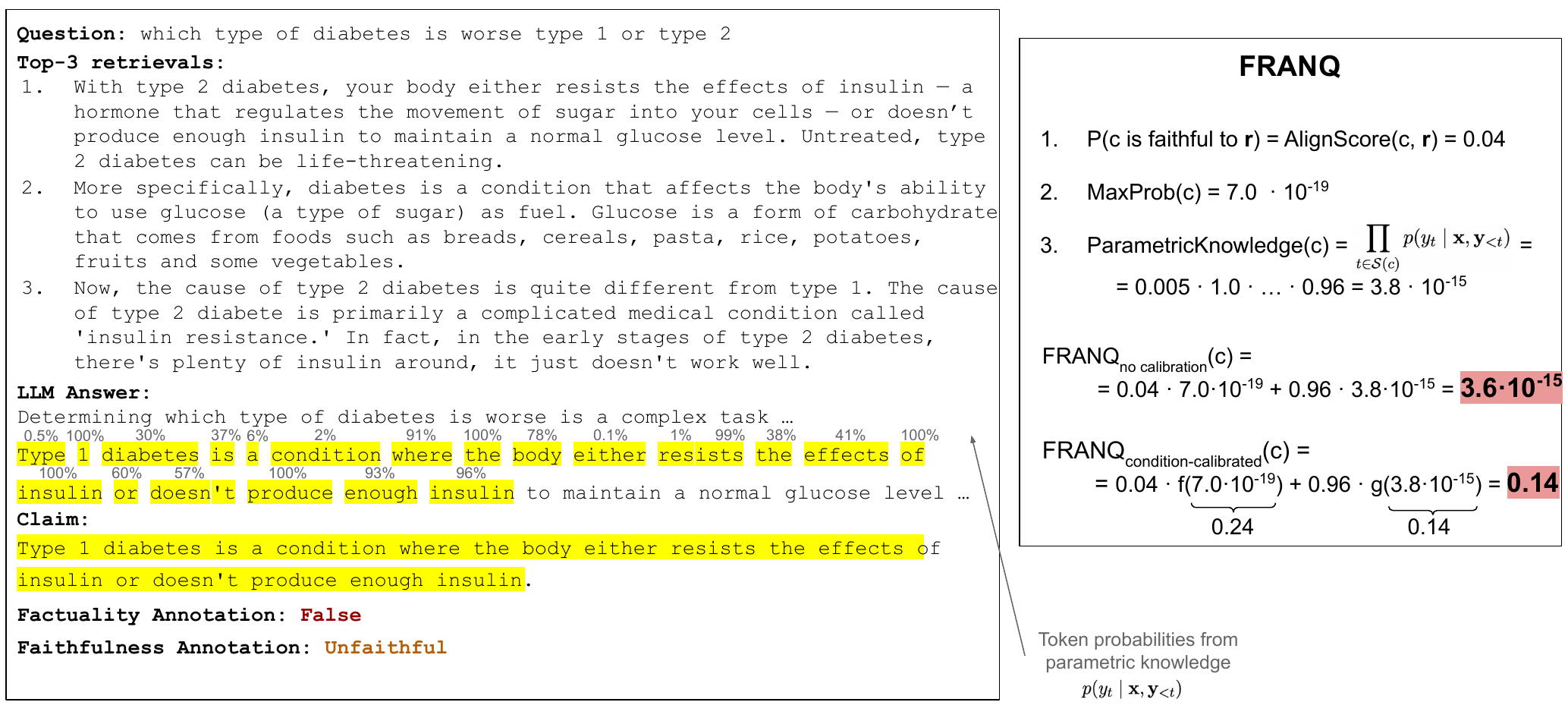}
        \caption{\textbf{Unfaithful–False.} \franq correctly identifies the claim as unfaithful and assigns a low factuality score using Parametric Knowledge, consistent across both the uncalibrated and calibrated versions.}
    \end{subfigure}
    \caption{Example outputs from \franq. \textit{Left}: Each example includes the input question, retrieved passages, the LLM-generated answer, a selected claim from the answer, and corresponding factuality and faithfulness annotations. Claims and their spans in the answer are highlighted in yellow. If a claim is faithful, its corresponding span in the retrieved passages is also highlighted. \textit{Right}: The \franq component scores and final factuality estimations, shown for both the uncalibrated and condition-calibrated versions.} 
    \label{fig:franq_examples}
  \end{figure*}

  In Figure~\ref{fig:franq_examples}, we demonstrate the behavior of \franq using three examples from a long-form QA dataset evaluated with \llama. We selected three representative claims and present their corresponding \franq scores for both the uncalibrated version and condition-calibrated version. The latest uses monotonic functions $f$ and $g$, fitted via isotonic regression for Claim Probability and Parametric Knowledge methods, respectively.

\section{The Usage of LLMs}
  In this study, large language models are examined primarily as the object of analysis. For practical tasks such as programming and writing, we also make limited use of LLM-based assistants (e.g., ChatGPT) for grammar correction and code debugging, with all such use carefully supervised by human researchers.

\end{document}